\documentclass[10pt,journal,compsoc]{IEEEtran}
\usepackage{cite}
\usepackage{multirow}
\usepackage{graphicx}
\usepackage{color}
\usepackage{dcolumn}
\usepackage{amsmath}
\usepackage{kantlipsum}
\usepackage{url}
\usepackage{microtype}
\usepackage{verbatim}
\usepackage{colortbl}
\usepackage{arydshln}
\usepackage{multirow}
\newcolumntype{K}[1]{>{\centering\arraybackslash}p{#1}}
\usepackage{booktabs}
\usepackage[flushleft]{threeparttable}
\usepackage{hhline}
\usepackage{algorithm}
\usepackage{algorithmic}
\usepackage{mathtools}
\renewcommand{\algorithmicrequire}{\textbf{Input:}}
\renewcommand{\algorithmicensure}{\textbf{Output:}}
\ifCLASSOPTIONcompsoc
\usepackage[caption=false, font=normalsize, labelfont=sf, textfont=sf]{subfig}
\else
\usepackage[caption=false, font=footnotesize]{subfig}
\fi
\begin{document}
\title{SCANN: Synthesis of Compact and Accurate Neural Networks}

\author{Shayan~Hassantabar,~Zeyu~Wang,~and~Niraj~K.~Jha,~\IEEEmembership{Fellow,~IEEE}
\thanks{This work was supported by IP Group and NSF Grant No. CNS-1617640 and CNS-1907381.
Shayan Hassantabar, Zeyu Wang, and Niraj K. Jha are with the Department
of Electrical Engineering, Princeton University, Princeton,
NJ, 08544 USA, e-mail:\{seyedh, zeyuwang,jha\}@princeton.edu.}}

\IEEEtitleabstractindextext{%
\begin{abstract}
Deep neural networks (DNNs) have become the driving force behind recent artificial intelligence 
(AI) research.  With the help of a vast amount of training data, neural networks can perform better 
than traditional machine learning algorithms in many applications.  An important problem with
implementing a neural network is the design of its architecture.  Typically, such an architecture is 
obtained manually by exploring its hyperparameter space and kept fixed during training. 
This approach is both time-consuming and inefficient.  Another issue is that modern neural networks
often contain millions of parameters, whereas many applications require small inference models due to 
imposed resource constraints, such as energy constraints on battery-operated devices.  However, 
efforts to migrate DNNs to such devices typically entail a significant loss of classification 
accuracy.  To address these challenges, we propose a two-step neural network synthesis methodology, 
called DR+SCANN, that combines two complementary approaches to design compact and accurate DNNs.    
At the core of our framework is the SCANN methodology that uses three basic architecture-changing 
operations, namely connection growth, neuron growth, and connection pruning, to synthesize 
feed-forward architectures with arbitrary structure.  These neural networks are not limited to the
multilayer perceptron structure.  SCANN encapsulates three synthesis methodologies that apply a 
repeated grow-and-prune paradigm to three architectural starting points.  DR+SCANN combines the SCANN methodology with dataset dimensionality reduction to alleviate the curse of dimensionality.    
We demonstrate the efficacy of SCANN and DR+SCANN on various image and non-image datasets. 
We evaluate SCANN on MNIST and ImageNet benchmarks.  Without any loss in accuracy, SCANN generates a 
46.3$\times$ smaller network than the LeNet-5 Caffe model. We also compare SCANN-synthesized 
networks with a state-of-the-art fully-connected feed-forward model for MNIST, and show $20\times$ 
($19.9\times$) reduction in number of parameters (floating-point operations) with little drop in 
accuracy.  On the ImageNet dataset, for the VGG-16 and MobileNetV2 architectures, we reduce the 
network parameters by $8.0\times$ and $1.3\times$ with a similar performance or improvement
over their respective baselines.  We also evaluate the efficacy of using dimensionality 
reduction alongside SCANN (DR+SCANN) on nine small to medium-size datasets. Using this 
methodology enables us to reduce the number of connections in the network by up to $5078.7\times$ 
(geometric mean: 82.1$\times$), with little to no drop in accuracy.  We also show that our synthesis 
methodology yields neural networks that are much better at navigating the accuracy vs. energy 
efficiency space.  This would enable neural network-based inference even on Internet-of-Things sensors.
\end{abstract}

}
\maketitle

\IEEEdisplaynontitleabstractindextext

\IEEEpeerreviewmaketitle

\IEEEraisesectionheading{\section{Introduction}\label{sec:introduction}}

\IEEEPARstart{A}{rtificial} neural networks have a long history, dating back to 1950's.  However, 
interest in neural networks has waxed and waned over the years.  The recent spurt in interest in deep
neural networks (DNNs) is due to large datasets becoming available, enabling them to be trained to
high accuracy. This trend is due to a significant increase in
computing power that speeds up the training process. DNNs demonstrate very
high classification accuracies for many applications of interest, e.g.,
image recognition, speech recognition, and machine translation.
They have become deeper, with tens to hundreds of layers. Thus, the
phrase `deep learning' is often associated with such 
DNNs. Deep learning refers to the ability of DNNs to learn
hierarchically, with complex features built upon simple ones.

The DNN architecture trained on a specific dataset has a great impact on the final performance of the 
model.  For example, Table~\ref{tab:imagenet-archs} compares several well-known DNNs designed for the 
ImageNet Large Scale Visual Recognition Challenge (ILSVRC) 2012-2016 \cite{deng2009imagenet}. 
We show the architectures in terms of the number of parameters in the network (\#params) and 
floating-point operations (FLOPs), as well as their performance on this task, i.e., the top-5 
accuracy.  Although all these well-known DNN architectures were obtained using the same training 
dataset and the same back-propagation (BP) algorithm for training weights, due to their architectural 
differences, their performance is vastly different in terms of classification accuracy, computational 
costs, and memory requirements. 

\begin{table}[t]
	\caption{Comparison of different classification models on ImageNet}
	\label{tab:imagenet-archs}
	\centering
	\begin{tabular}{lcccc}
		\toprule
		Architecture     & Top-5 accuracy &  \#Params & FLOPs \\
		\midrule
		AlexNet \cite{krizhevsky2009learning} & $84.7\%$ & 62M  & 1.5B \\
		VGG-16 \cite{DBLP:journals/corr/SimonyanZ14a} & $92.3\%$ & 138M & 30.9B \\
		Inception \cite{szegedy2015going} & $93.3\%$ & 6.4M & 2B \\
		ResNet-152 \cite{he2016deep} & $95.5\%$ & 60.3M & 11B \\
		MobileNetV2 \cite{sandler2018mobilenetv2} & $91.0\%$ & 3.5M & 300M \\
 		\bottomrule
	\end{tabular}
\end{table}

Though critically important, how to derive an appropriate DNN architecture for small, medium, and 
large datasets has remained a vexing problem.  Since the DNN architecture directly influences the 
learned representations and thus the performance of the model, this is an important challenge in 
deploying DNNs in practice and using their knowledge distillation power in various applications, such 
as smart healthcare \cite{hassantabar2020coviddeep, hassantabar2020diagnosis}.  Typically, it takes 
researchers a huge amount of time through much trial-and-error to find a good architecture
because the search space is exponentially large with respect to many
of its hyperparameters. Furthermore, these architectures need to be trained on large datasets. 
This approach suffers from four major problems: 

\begin{itemize}
    \item \emph{Fixed network architecture}: These methods use the BP algorithm to train the weights, 
and not to optimize the architecture. This means that the DNN architecture, including the depth and 
connections of the network, is kept fixed during the training process.  This does not lead to better 
DNN architectures.
    \item \emph{Lengthy search process}: Searching for an accurate DNN architecture through 
trial and error is inefficient. This problem is exacerbated when the DNN becomes larger.  Each trial 
can easily take tens of hours on fast graphical processing units (GPUs). In addition, it takes 
months to design more efficient architectures for certain tasks, such as the architectures 
shown in Table \ref{tab:imagenet-archs} for image classification. 
    \item \emph{Architectural redundancy}: Most DNNs suffer from substantial storage and computation 
redundancy \cite{han2016eie}. For example, Dai et al.~\cite{dai2019nest} show that the number of 
parameters and the FLOPs of ResNet-50 can be reduced by 4.1$\times$ and 2.1$\times$, respectively, 
without loss of accuracy.
    \item \emph{Need for large datasets}: Collecting a large number of data instances and manually 
labeling them is a costly process, especially in domains where experts are needed to label the data 
instances, such as data collected for healthcare \cite{hassantabar2021mhdeep}.  Using synthetic data 
generated from the same distribution as the real data, however, can reduce the need for large 
datasets \cite{hassantabar2020tutor}.
\end{itemize}


As the number of features, i.e., dimension, of the dataset increases, in order to generalize 
accurately, we need exponentially more data.  This is another challenge that is referred to as  
the \emph{curse of dimensionality}.  Hence, one way to reduce the need for large amounts of data is 
to reduce the dimensionality of the dataset.  In addition, with the same amount of data, by reducing 
the number of features, the accuracy of the inference model may also improve to a degree. 
However, beyond a certain point, which is dataset-dependent, reducing the number of features may lead 
to loss of information, which may lead to inferior classification results.  

To address the above problems, we propose a new DNN synthesis tool called DR+SCANN that combines two 
different approaches to synthesize very compact, yet accurate, DNNs.  DR+SCANN starts the 
DNN synthesis from a seed DNN architecture.  We refer to this architecture as the baseline model. 
The baseline model can be chosen from among the well-known architectures for certain datasets,
e.g., ImageNet, or a well-performing fully-connected (FC) architecture.  First, we use
dimensionality reduction (DR) methods for the non-image datasets to reduce their feature size. 
This helps with network compression while improving its classification accuracy. 

The second step of our methodology is its main part, called SCANN. SCANN starts DNN synthesis 
with a seed architecture. It uses three architecture-changing operations in multiple iterations 
to synthesize accurate and compact models.  It focuses on the FC layers of the architecture and allows 
DNNs to grow connections and neurons based on the gradient information so that the model can be 
adapted to the task at hand.  SCANN uses two different operations for network growth, namely 
connection growth and neuron growth. Then, SCANN prunes away insignificant connections in the 
architecture based on the magnitude information.  Unlike previous grow-and-prune synthesis 
approaches \cite{dai2019nest}, SCANN does not limit the architecture to the multi-layer perceptron 
(MLP) structure.  By allowing any neuron to connect to any other neuron in the DNN architecture, 
SCANN allows skipped connections in the network. In addition, although previous grow-and-prune 
synthesis approaches allow the weights and connections to be learned during the training process, they 
do not allow a change in the number of artchitecture layers during training.  SCANN removes this 
limitation, allowing it to derive better architectures.

We use SCANN and DR+SCANN to synthesize various compact DNNs for small, medium, and large datasets. 
We used the SCANN methodology to generate compact DNNs for MNIST \cite{lecun1998gradient} and 
ImageNet \cite{deng2009imagenet} datasets.  We use DR+SCANN to generate compact DNNs for several 
non-image datasets.  As we show later, DR+SCANN leads to drastic reductions in the number of 
parameters and computational cost of the model relative to the FC DNN baselines while also improving 
classification performance. 

The major contributions of this work can be summarized as follows: 

\begin{itemize}
    \item We present SCANN, a grow-and-prune synthesis methodology, that yields compact and accurate 
feed-forward neural networks for datasets spanning small to large sizes.  SCANN addresses a
limitation of prior work that fixes the number of layers in the architecture prior 
to the training process. 
    \item We use DR methods to mitigate the curse of dimensionality and improve the performance while 
compressing the network architecture. 
    \item We propose a two-step DNN synthesis process, DR+SCANN, that combines DR with SCANN
to learn very compact and accurate neural network models.
    \item We evaluate the performance of SCANN on MNIST and ImageNet datasets with various seed 
architectures. SCANN targets the FC layers of image-based architectures since these layers
contain a large fraction of all parameters.
    \item We evaluate the performance of DR+SCANN on nine small to medium datasets and demonstrate 
$1.2\times$ to $5078.7\times$ compression in network parameters with little to no drop in accuracy. 
We demonstrate that DR+SCANN yields DNNs that are very energy-efficient, while offering a similar 
accuracy to other methods.  This opens the door for such DNNs to even be used in Internet-of-Things 
(IoT) sensors.  
\end{itemize}


The rest of the article is organized as follows.  Section \ref{sect:related} describes
related work. Section \ref{sect:method} describes the SCANN and DR+SCANN synthesis 
methodologies in detail.  Section \ref{sect:expresults} provides results of our evaluations. 
Section \ref{sect:discussion} provides a short discussion.  Finally, Section
\ref{sect:conclusion} concludes the article.

\section{Related Work}
\label{sect:related}

In this section, we review some of the previous work in two related areas:
DR and automatic architecture synthesis.

\subsection{Dimensionality Reduction}
The high dimensionality of many datasets used in various applications of machine learning leads
to the curse of dimensionality problem. Therefore, researchers have explored DR methods to improve the performance of machine learning models by decreasing the number of features. 
Traditional DR methods include Principal Component Analysis (PCA), Kernel PCA, 
Factor Analysis (FA), Independent Component Analysis (ICA), as well as Spectral Embedding methods. 
Some graph-based methods include Isomap \cite{tenenbaum2000global} and Maximum Variance Unfolding 
\cite{weinberger2006introduction}.  FeatureNet \cite{bhardwaj2018dimensionality} uses community 
detection in small sample size datasets to map high-dimensional data to lower dimensions. Other 
DR methods include stochastic proximity embedding, linear discriminant analysis , and t-distributed 
stochastic neighbor embedding \cite{maaten2008visualizing}.  A detailed survey of DR methods can be
found in \cite{van2009dimensionality}.

\subsection{Automatic Architecture Synthesis}
There are three different categories of automatic architecture synthesis
methods that have been proposed by researchers: evolutionary algorithm,
reinforcement learning algorithm, and structure adaptation algorithm.

\subsubsection{Reinforcement Learning Algorithm}
In a recent trend, reinforcement learning (RL) has been used to search for architectures in an 
automated flow \cite{DBLP:conf/iclr/ZophL17}.  This approach is known as neural architecture search 
(NAS).  A typical NAS framework uses a controller based on recurrent neural networks to iteratively 
generate candidate architectures in the search process.  Based on the performance of the candidate 
architectures, the RL controller gets updated in the next iteration.  Zoph and Le 
\cite{DBLP:conf/iclr/ZophL17} use a recurrent neural network as a controller to generate a string 
that specifies the network architecture.  They use the performance of the generated network on a
validation set as the reward signal to compute the policy gradient and update the controller. 
NASNet \cite{DBLP:conf/cvpr/ZophVSL18} yields a new search space that is transferable. It uses RL 
to find the best convolutional layers on the CIFAR-10 dataset and then uses these layers 
for the ImageNet dataset by stacking multiple copies each with its own parameters. 
RL-based approaches can also be used to design efficient DNN architectures for mobile platforms. 
MNasNet \cite{tan2019mnasnet} uses this approach to achieve top-1 accuracy of $75.2\%$ on the 
ImageNet classification task with very low latency on the mobile platforms.  Although RL-based 
architecture search approaches have been successful, this process remains computationally intensive. 

\subsubsection{Evolutionary Algorithm}
The use of an evolutionary algorithm to select a DNN architecture dates back to 1989 
\cite{miller1989designing}.  One of the seminal works in neuroevolution is the NEAT algorithm 
\cite{stanley2002evolving}, which uses direct encoding of every neuron and connection to 
simultaneously evolve the network architecture and weights through weight mutation, connection 
mutation, node mutation, and crossover.  Recent years have seen extensions of the evolutionary 
algorithm to generate convolutional neural networks (CNNs).  For example, Xie and Yuille 
\cite{xie2017genetic} use a concise binary
representation of network connections and demonstrate a comparable
classification accuracy to previous human-designed architectures.
It is also beneficial to combine efficient evolutionary search with various performance predictors 
to optimize architectural hyperparameters \cite{dai2018chamnet, hassantabar2019steerage}. 
FBNetV$3$ \cite{dai2020fbnetv3} adds the training recipe (i.e., training hyperparameters) to the 
evolutionary search process. As a result, the search process can find higher accuracy-recipe 
combinations.

\subsubsection{Structure Adaptation Algorithm}
Several previous works achieve compact and accurate neural networks through structure adaptation 
algorithms.  One such method is network pruning, which has been used in several works 
\cite{han2015learning, dai2019nest,  yang2016designing, yang2018netadapt}.  Structure adaptation 
algorithms can be constructive or destructive.  Constructive algorithms start with
a small neural network and grow it into a larger more accurate neural network.  Destructive 
algorithms start with a large neural network and prune connections and neurons to get rid of 
redundancy while maintaining accuracy.  NeST \cite{dai2019nest} is a network synthesis tool that 
combines both the constructive and destructive approaches in a grow-and-prune synthesis paradigm. 
However, its limitation is that both growth and pruning are performed at a specific DNN layer. 
Thus, network depth cannot be adjusted and is fixed throughout training.  In the next section, we 
show this problem can be solved by synthesizing a general feed-forward network instead of an MLP 
architecture, allowing the DNN depth to be learned dynamically during the training process.

Several works have also proposed more efficient building blocks for CNN architectures 
\cite{wu2017shift, DBLP:conf/cvpr/ZhangZLS18, sandler2018mobilenetv2}.  They result 
in compact networks, with much fewer parameters, while maintaining or improving performance. 
Platform-aware search for an optimized DNN architecture has also been used in this area.
Yin et al.~\cite{yin2019hardware} combine the grow-prune synthesis methodology with
hardware-guided training to achieve compact long short-term memory cells. 

Orthogonal to the above works, quantization has also been used to reduce 
computations in a network with little to no accuracy drop \cite{
zhu2016trained}.

\section{Methodology}
\label{sect:method}
In this section, we describe various parts of the proposed DNN synthesis methodology.  First, we give 
an overview of our two-step DNN synthesis approach.  Then, we discuss the SCANN synthesis methodology 
to learn both the weights and an efficient DNN architecture.  We then explain our DR pre-processing 
step to not only reduce the number of features, but to improve the classification accuracy as well. 

\subsection{Framework Overview}
\label{sect:overview}
\begin{figure*}[]
	\centering
	\includegraphics[width=6.5in]{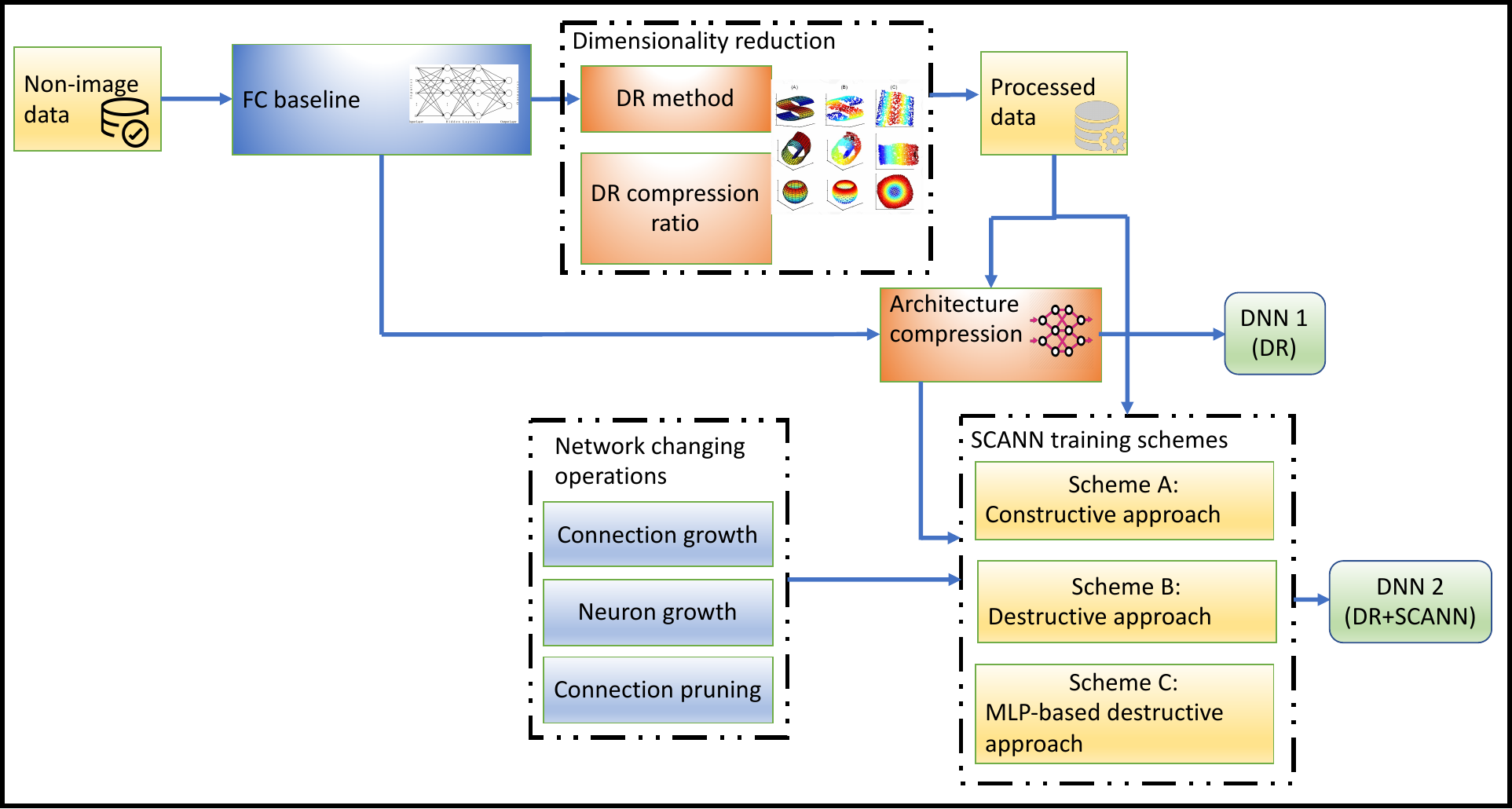}
	\caption{Block diagram of DR+SCANN.  DR enables a smaller dataset to be used to
    synthesize the DNN 1 architecture and the DR+SCANN architecture using SCANN. The SCANN step uses 
three architecture-changing operations in three different training schemes and uses the
reduced-dimension dataset. }
	\label{fig:dr-scann-block-diagram}
\end{figure*}

\begin{figure}[]
	\centering
	\includegraphics[width=3.61in]{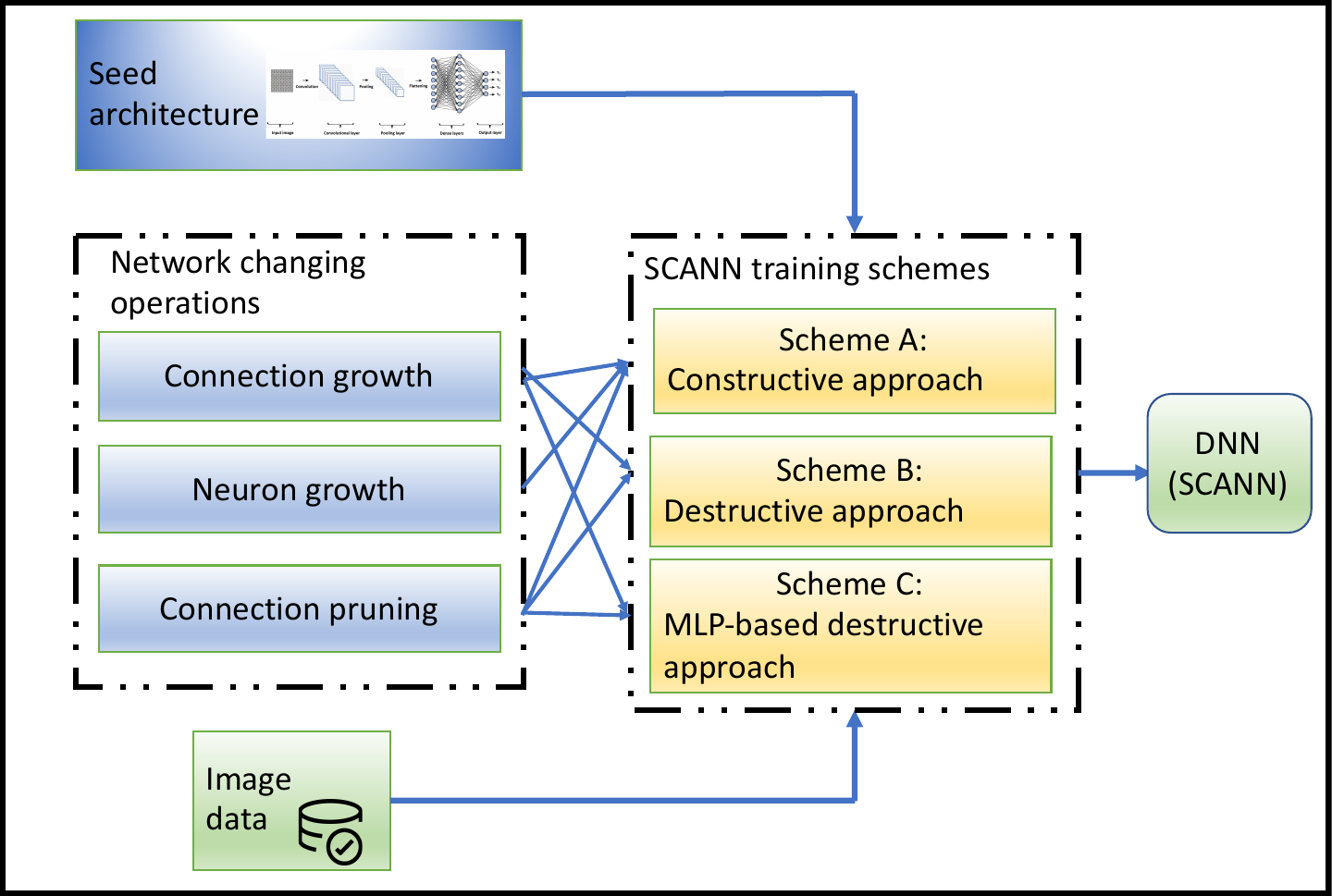}
	\caption{Block diagram of SCANN. SCANN uses three different architecture changing operation 
    in its three training schemes to change the baseline seed architecture and synthesize compact architectures. }
	\label{fig:scann-block-diagram}
\end{figure}

Our DNN synthesis methodology covers both non-image and image datasets. 
For the non-image datasets, we use a two-step sequential method, which we refer to as DR+SCANN. 
We illustrate the block diagram of DR+SCANN framework in Fig.~\ref{fig:dr-scann-block-diagram}. 
For the image dataset, we use just the SCANN DNN synthesis framework, as shown in 
Fig.~\ref{fig:scann-block-diagram}.  In the DR step, we first modify the dataset by conducting 
dataset normalization and performing DR on dataset features.  DR is aimed at alleviating the curse of
dimensionality and increasing classification accuracy. As a result, we also obtain a smaller DNN 
architecture. 

For the non-image datasets, we compare the DNN models designed by the DR+SCANN methodology with 
FC baselines obtained by training various DNNs (with 
different numbers of layers and different numbers of neurons per layer) and verifying their 
performance on the validation set.  The DR step chooses the best DR method and feature 
compression ratio for each dataset.  We also demonstrate that we can use a smaller FC baseline 
architecture and still improve its classification accuracy when the number of features is reduced. 
The process of neural network compression with DR is explained in Section~\ref{sect:dr}.

For the image datasets, MNIST and ImageNet in our experiments, we use well-known DNN models as the 
seed architectures.  We use the SCANN synthesis methodology for designing efficient DNNs. 
SCANN combines gradient information to grow connections, activation information to grow neurons, and 
magnitude information to prune insignificant connections in the FC layers of the network.  By allowing 
skipped connection between neurons in the architecture, SCANN addresses the limitation of prior work 
that requires the DNN depth to be fixed prior to the training phase.  These architecture-changing 
operations are used in three different training schemes.  This process is explained in 
Section \ref{sect:scann}. 

\subsection{Dimensionality reduction}
\label{sect:dr}
In this step, we first normalize the data. Data normalization generally leads to higher accuracy and better noise tolerance. We use range normalization for this purpose. 

$$
x_{normalized} = \frac{x - min (x)}{max(x) - min(x)}
$$

\noindent
This scales each input data instance into the [0,1] range.  Next, we use DR to reduce the number of 
features in the dataset.  An $N \times d$-dimensional dataset is mapped onto an 
$N \times k$-dimensional space, $k < d$, using various DR methods.  We explore nine such methods, 
including four random projection (RP) methods. 

Dimensionality reduction with RP is based on the Johnson-Lindenstrauss 
lemma \cite{sivakumar2002algorithmic, dasgupta2003elementary}.  The essence of this lemma is that 
sufficiently high-dimensionality data points can be projected onto a suitable lower 
dimension, while approximately maintaining inter-point distances.  More 
precisely, this lemma shows that the distance between the points changes only 
by a factor of $(1 \pm \varepsilon)$, when they are randomly projected onto 
the subspace of $\mathcal{O}(\log{} \frac{d}{\varepsilon^2})$ dimensions, for
any $0 < \varepsilon < 1$.  

RP uses a projection matrix to compute the features in the lower dimension.
The RP matrix $\Phi$ can be generated in several ways.  Here, we discuss four
RP matrices that we used in our implementation. 
One approach is to generate $\Phi$ using a Gaussian distribution. 
In this case, the entries $\phi_{i,j}$ are i.i.d. samples drawn from a Gaussian distribution 
$\mathcal{N}(0,\frac{1}{k})$. Another RP matrix can be obtained by sampling entries from 
$\mathcal{N}(0,1)$.  These entries are shown below.
$$
\phi^{1}_{ij} \sim \mathcal{N}(0,\frac{1}{k}) \qquad \phi^{2}_{ij} \sim \mathcal{N}(0,1)
$$

Achlioptas \cite{achlioptas2001database} proposed several other sparse RP matrices. Two of these 
proposals are as follows, where entries $\phi_{ij}$'s are independent random variables that are 
drawn based on the following probability distributions: 

$$
\phi^{3}_{ij} = \begin{cases}
+1 & \text{with probability   }  \frac{1}{2}\\
-1 & \text{with probability   } \frac{1}{2}\\
\end{cases}
$$

$$
    \phi^{4}_{ij} = \sqrt{\frac{3}{k}} \begin{cases}
    1 & \text{with probability  } \frac{1}{6}\\
    0 & \text{with probability  } \frac{2}{3}\\
    -1 & \text{with probability  } \frac{1}{6}\\
    \end{cases}
$$

The other DR methods that we use are PCA, FA, Isomap, ICA, and Spectral Embedding. Implementations of 
these methods are obtained from the Scikit-learn machine learning library \cite{pedregosa2011scikit}.

\begin{figure}[]
	\centering
	\includegraphics[width=3.5in]{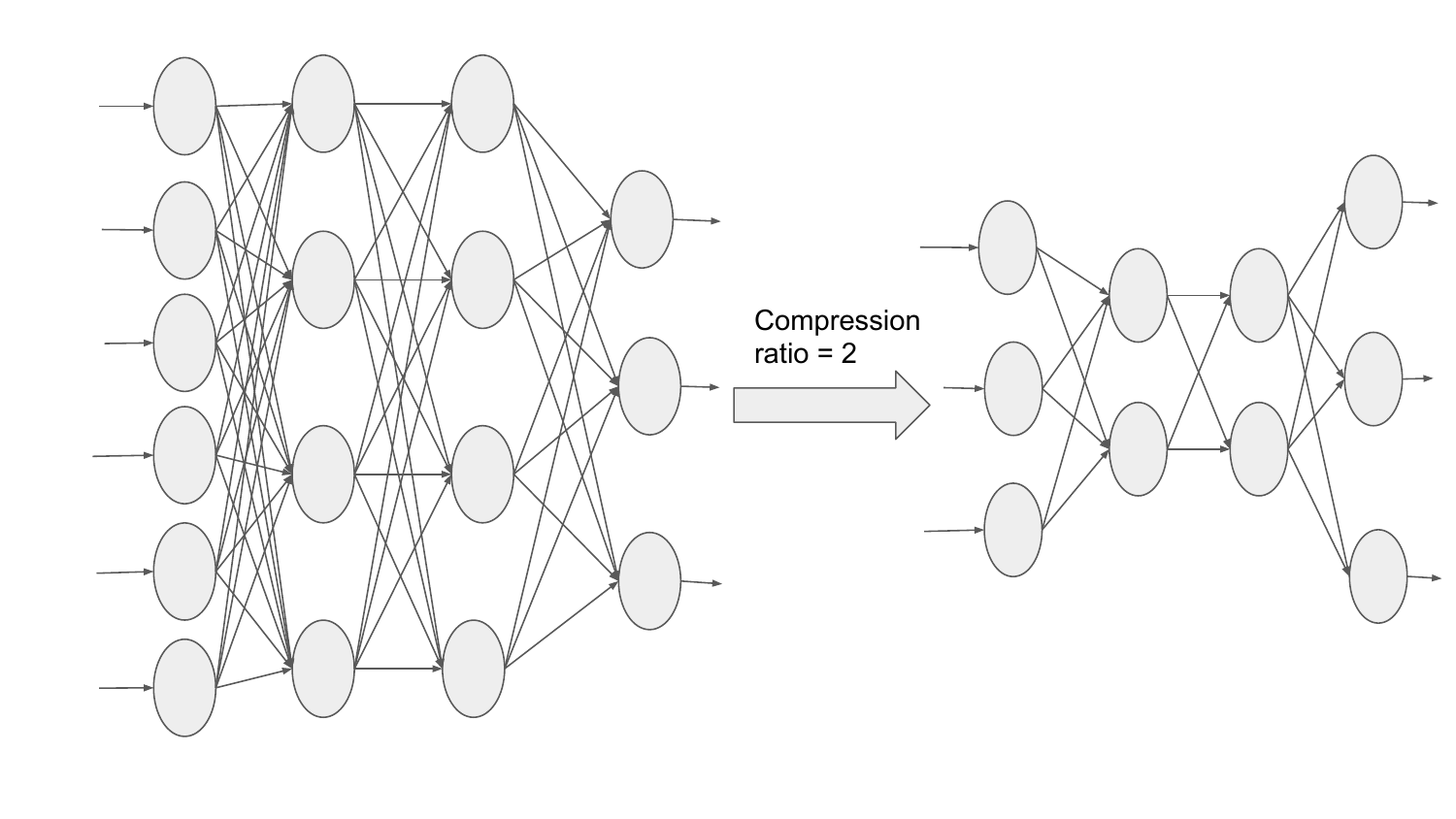}
	\caption{Compressing the neural network by a 2$\times$ compression ratio: the number of 
neurons in each layer, except the last layer, is reduced by a factor of 2.}
	\label{fig:NNcompression}
\end{figure}

DR maps the dataset into a vector space of lower dimension.  As the number of features reduces, the 
number of neurons in the input layer of the neural network decreases accordingly.
However, since the dataset dimension is reduced, one might expect the task of classification to 
become easier.  This means that we may be able to use a smaller DNN architecture, in general. 
We show that we can indeed reduce the number of neurons in all layers, not just the input layer.
In fact, we show that we can use a DNN architecture with the number of neurons in each layer 
reduced by the same \emph{feature compression ratio} obtained in the DR step, except for the output 
layer.  We use this ratio to show that DR can increase classification accuracy while enabling the use 
of a smaller DNN architecture.  Fig.~\ref{fig:NNcompression} shows an example of the process of 
reducing the number of neurons in the architecture.  We refer to this model as the \emph{DR} model.

Algorithm \ref{alg:dr} summarizes the process of dataset DR and architecture compression. 
We obtain
the reduced-dimension dataset for all the DR methods and various feature compression ratios. 
Dimensionality reduction methods perform differently on various datasets.  We also use early stopping 
to terminate DR methods that do not perform well on the validation set. 
Furthermore, we stop reducing the DR compression ratio when the performance drops significantly.
As we reduce the number of features, we reduce the number of neurons in each layer of the initial 
architecture with the same ratio.  Note that the number of neurons in each layer could be
reduced using different ratios. However, the combinatorial explosion of choices makes navigation
of this larger search space computationally prohibitive.  We show later that uniform reduction
across layers works very well in practice.  Then, we train the new architecture using the 
reduced-dimension training set and evaluate it on the reduced-dimension validation set.  Finally, 
we select the architecture with the highest validation accuracy and record its test accuracy. 
The output of this algorithm is the best performing architecture $\hat{A}$ (on the validation
dataset), its corresponding test accuracy, and the corresponding reduced-dimension dataset.

\begin{algorithm}[]
\caption{Dimensionality reduction process}
\label{alg:dr}
\begin{algorithmic}[1]
\renewcommand{\algorithmicrequire}{\textbf{Input:}}
\renewcommand{\algorithmicensure}{\textbf{Output:}}
\REQUIRE Original dataset (training, validation, and test set) of size $N$ with $d$ features, 
initial network architecture $\mathcal{A}_{init}$, with $m$ layers of $|\mathcal{L}_1|, |\mathcal{L}_2|, \dotsc ,|\mathcal{L}_m|$ neurons

\ENSURE Reduced-dimension dataset $\Tilde{X}$, compressed architecture $\hat{A}$, and test accuracy

\STATE Dataset min-max normalization 
\FORALL{DR methods}
\FORALL{feature compression ratios}

\STATE Reduce the dataset dimension using the DR method to get $\Tilde{X}_{train}$, $\Tilde{X}_{validation}$, and 
$\Tilde{X}_{test}$


\FOR{$i=1:m-1$}
\STATE $ |\mathcal{\hat{L}}_i| =\lfloor \frac{|\mathcal{L}_i|}{\textit{feature compression ratio}} \rfloor $\
\STATE Train the new architecture $\hat{A}$ with ($\Tilde{X}_{train}, y_{train}$) 
\STATE Evaluate the trained architecture $\hat{A}$ on ($\Tilde{X}_{validation}, y_{validation}$) 
\STATE Save the trained architecture with the highest validation accuracy

\ENDFOR

\ENDFOR
\ENDFOR

\STATE $testAcc$ = Accuracy of the trained architecture $\hat{A}$ on ($\Tilde{X}_{test}, y_{test}$) 
\RETURN $\Tilde{X}$, $\hat{A}$,  and $testAcc$
\end{algorithmic}
\end{algorithm}

\subsection{SCANN Synthesis Methodology}
\label{sect:scann}

Next, we explain the SCANN methodology that leverages both destructive and constructive architecture 
synthesis approaches through a grow-and-prune synthesis paradigm.  As a result, the synthesis cost of 
this process is significantly reduced compared to RL-based approaches.  SCANN can also be used in 
conjunction with the DR process, as explained in Section \ref{sect:dr}.  DR+SCANN works on the 
reduced-dimension dataset whereas SCANN works on the original dataset. 

We first propose a technique to address the limitation of prior work that requires the number of 
layers of DNN architecture to be fixed prior to the training process.  Then, we introduce three
basic architecture-changing techniques that enable the synthesis of an optimized feed-forward network 
architecture.  
Finally, we describe three 
training Schemes, A, B, and C, that can be used to learn 
the weights and connections in the network during the training process.  Each of these training 
schemes uses a different approach to synthesizing efficient DNN architectures.  Scheme A is a 
constructive approach that starts from a small network and iteratively grows the network to a larger 
one.  On the other hand, Schemes B and C are based on destructive synthesis that starts from a 
larger network and iteratively prunes the architecture to a smaller one. 

\subsection{Depth Change}

\begin{figure}[!h]
	\centering
	\subfloat[]{\includegraphics[scale=0.35]{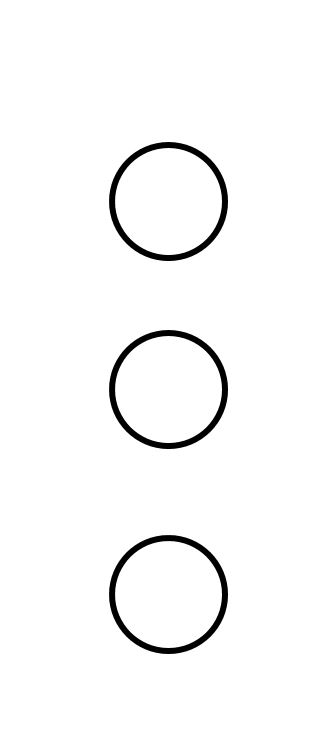}}
	\label{fig:1a}
	\hfil
	\subfloat[]{\includegraphics[scale=0.35]{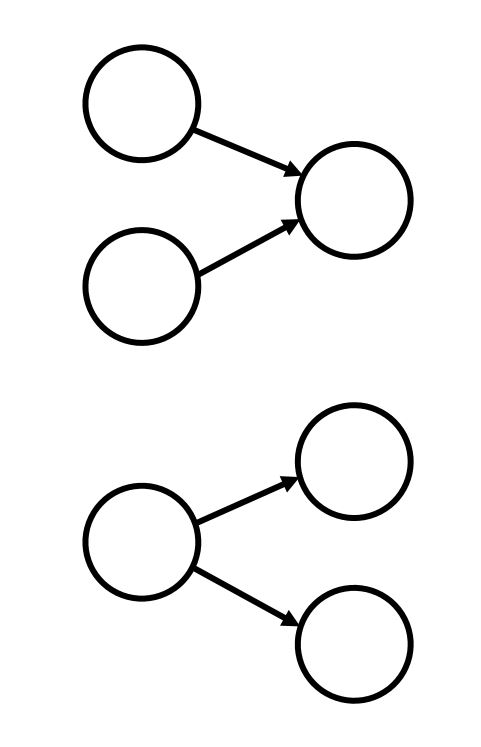}}
	\label{fig:1b}
	\subfloat[]{\includegraphics[scale=0.35]{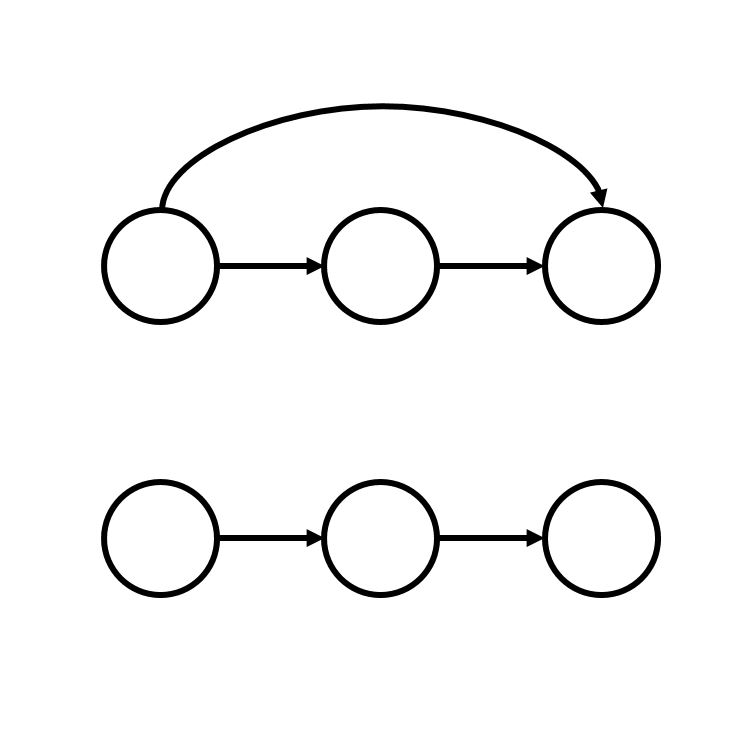}}
	\label{fig:1c}
	\caption{Connection pattern determines network depth. Only hidden neurons
		are shown. (a) One hidden layer, (b) two hidden layers, and (c) three hidden
		layers.}
	\label{fig:pattern}
\end{figure}

To address the problem of having to fix the depth of the DNN (number of layers in the architecture) 
prior to the training process, we adopt a general feed-forward architecture instead of an MLP 
structure.  Specifically, in this setting, depth is determined by how hidden neurons are connected 
and thus can be changed through the rewiring of hidden neurons. As shown in Fig.~\ref{fig:pattern}, 
depending on how the hidden neurons are connected, they can form one, two, or three hidden layers.
In addition, by allowing skipped connections in the architecture, we address the limitation of 
MLP structures in learning the architecture during the training process. 

\subsection{SCANN: Overall Workflow}
The overall workflow for architecture synthesis is shown in Algorithm \ref{alg:scann}. The
synthesis process iteratively alternates between architecture change and weight training. 
Thus, the network architecture evolves along the way.  The growth phase uses gradient information 
to gradually grow connections and activation information to grow neurons, in order to achieve the 
desired accuracy.  In the pruning phase, the magnitude information is used to remove the redundant 
connections.  After a specified number of iterations, the checkpoint that achieves the best 
performance on the validation set is output as the final network.  

\begin{figure}[]
	\centering
	\includegraphics[width=3.5in]{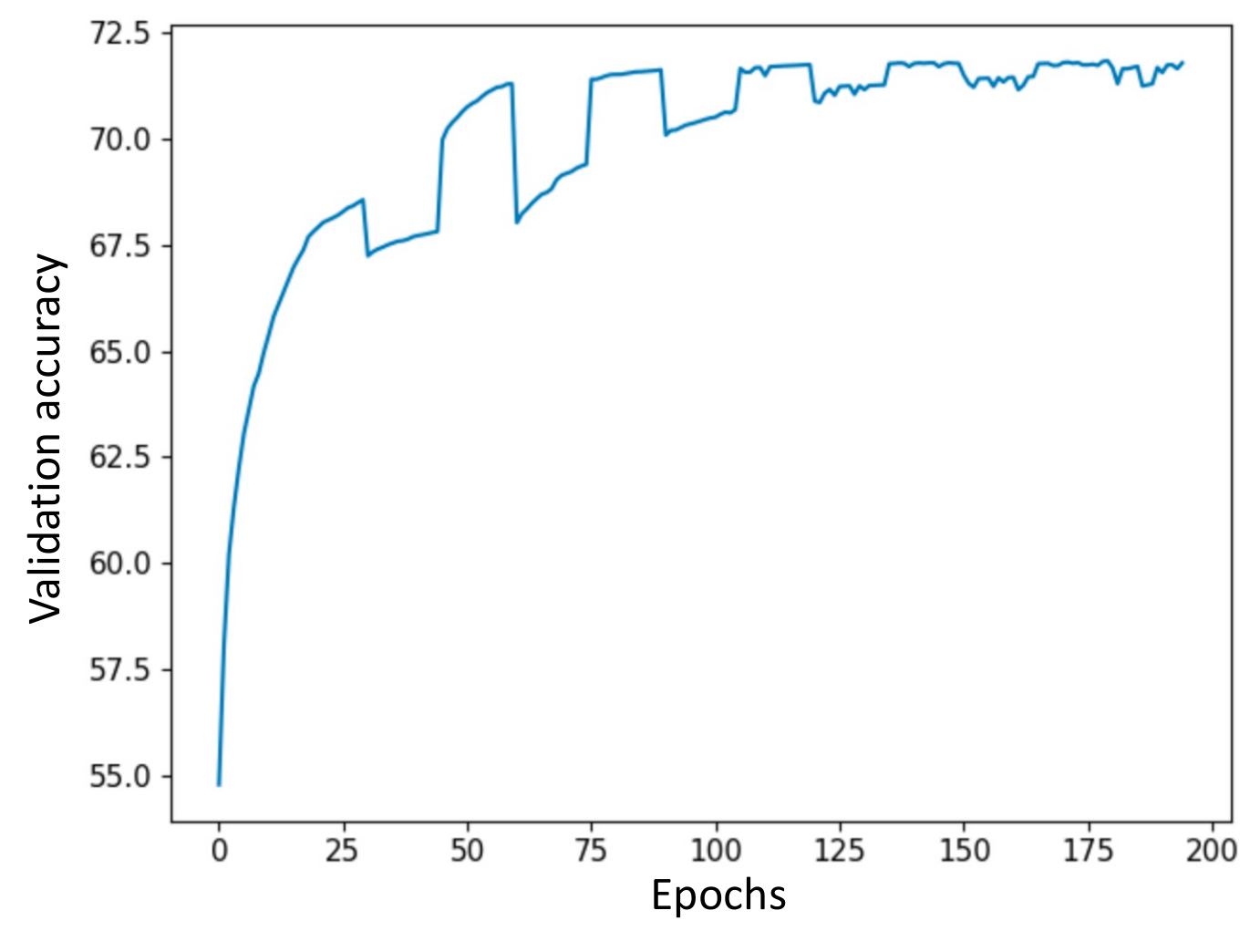}
	\caption{The impact of iterative grow-and-prune process on recovering the lost accuracy. }
	\label{fig:gp-mobilenetv2}
\end{figure}

Next, we first elaborate on the three basic architecture-changing operations and then introduce 
three different training schemes based on how the architectures evolve. 
The process of applying 
architecture-changing operations in the flow of Algorithm \ref{alg:scann} differs in each training 
scheme.  
In general, we found that using iterative growth-and-pruning enables both higher 
accuracies and compression ratios.  Fig.~\ref{fig:gp-mobilenetv2} shows the accuracy versus 
training epochs when applying SCANN to the MobileNetV2 architecture for the ImageNet dataset. 
As can be seen, pruning leads to a drop in accuracy after the growth operation. However, applying 
growth and pruning over multiple iterations enables the architecture to recover from the loss in 
performance. 

\begin{algorithm}[h]
	\caption{Automatic architecture synthesis}
	\label{alg:scann}
	\begin{algorithmic}[l]
		\REQUIRE Initial network architecture $A_{init}$, weights $W_{init}$, and
		maximum number of iterations $I_{max}$
		
		\WHILE{maximum iterations $I_{max}$ not reached}
		\STATE {\textbf{(a)} Perform one of the three basic architecture-changing
			operations (different in various schemes)\;}
		\STATE {\textbf{(b)} Train weights of the network and test its performance on the validation set\;}
		\ENDWHILE
		
		\ENSURE Final network architecture $A_{final}$ and associated weights
		$W_{final}$ that achieve the best performance on the validation set
		
	\end{algorithmic}
\end{algorithm}

\subsection{Basic Architecture-changing Operations}
\begin{figure*}[!ht]
    \centering
    \includegraphics[scale=0.45]{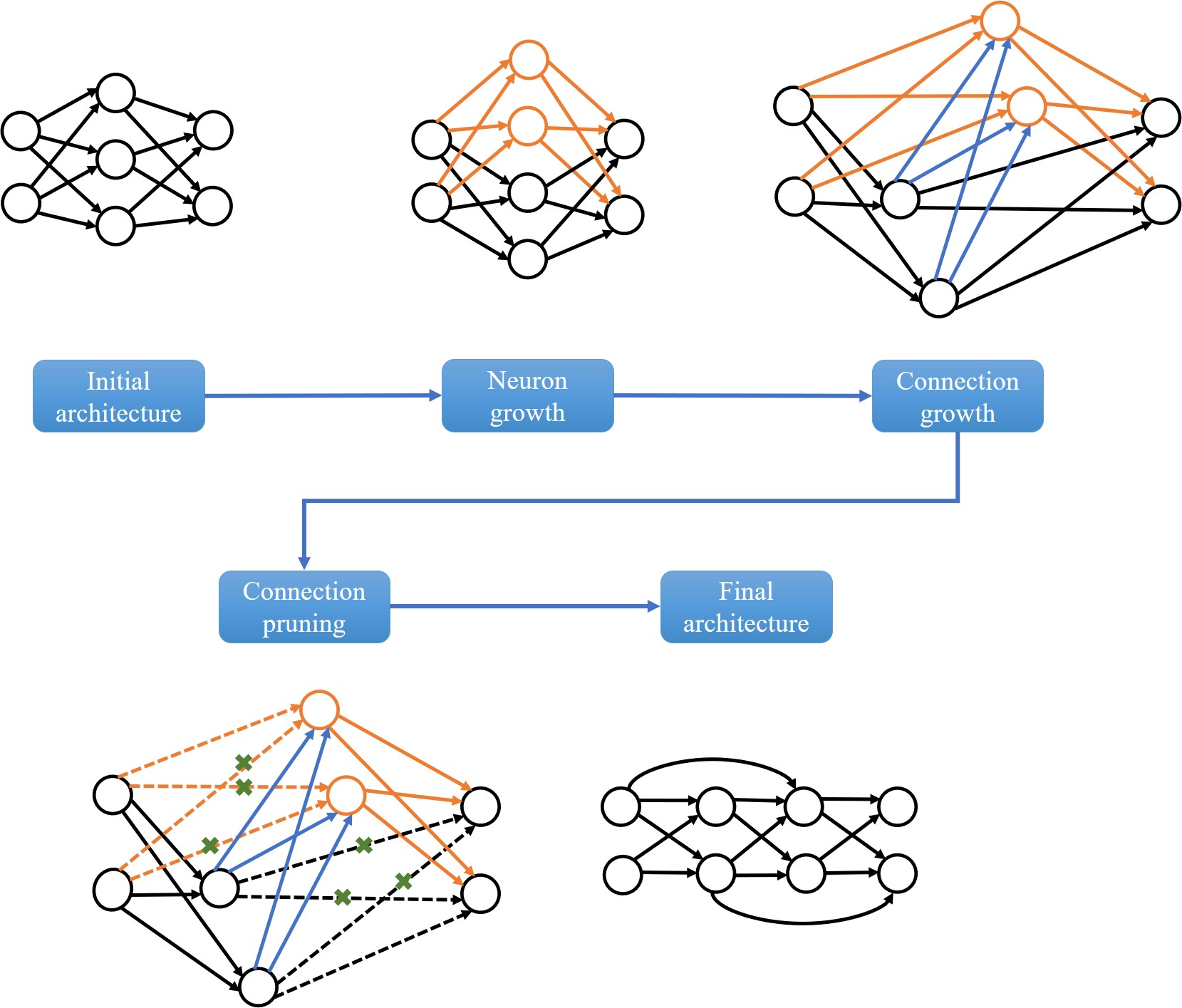}
    \caption{An MLP architecture with one hidden layer evolves into a
non-MLP architecture with two hidden layers through a sequence of neuron
growth, connection growth, and connection pruning.}
\label{fig:scheme}
\end{figure*}

Three basic operations, connection growth, neuron growth, and connection pruning, are used to 
evolve a feed-forward network architecture through multiple iterations.  Fig.~\ref{fig:scheme} shows 
a simple example in which an MLP architecture with one hidden layer evolves into a
non-MLP architecture with two hidden layers, with a sequence of basic operations mentioned above. 

Next, we describe these three operations. We denote the $i$th hidden neuron
as $n_i$, its activity as $x_i$, and its preactivity as $u_i$, where
$x_i = f(u_i)$ and $f$ is the nonlinear activation function. We denote the depth of
$n_i$ by $D_i$ and the loss function by $L$. Finally, we denote the connection
between $n_i$ and $n_j$, where $D_i < D_j$, as $w_{ij}$. In our implementation, we use a mask-based approach to ignore the dormant connections.

\subsubsection{Connection Growth}
The connection growth algorithm greedily adds connections between neurons that are unconnected. 
The initial weights of all newly added connections are set to 0.  Depending on how connections are 
added, we use two different methods, as shown in Algorithm~\ref{alg:2}.

\begin{itemize}
\item{\textbf{Gradient-based growth:}}
Gradient-based growth was proposed by Dai et al.~\cite{dai2019nest}. 
It adds connections that tend to reduce the loss function $L$ significantly. 
Suppose two neurons $n_i$ and $n_j$ are not connected and
$D_i \leq D_j$, then gradient-based growth adds a new connection $w_{ij}$
if $\left|\frac{\partial L}{\partial u_j}x_i\right|$ is large.
We evaluate the $\left| \partial L / \partial w \right|$ for all the dormant connections $w$ and 
activate the ones with the largest values.  This is done based on a large data batch $B$. 
We use a predefined threshold to activate the dormant neurons. This threshold can be chosen based on a certain percentage of elements in the computed gradient matrix.

The intuition behind this approach is the Hebbian theory
that states "neurons that wire together fire together" \cite{lowel1992selection}. The connections activated based on this theory would have a strong correlation between presynaptic and postsynaptic cells. Therefore, this translates to the large $\left|\frac{\partial L}{\partial u_j}x_i\right|$ values. 

\item{\textbf{Full growth:}}
Full growth restores all possible connections to the network.


\begin{algorithm}[h]
	\caption{Connection growth algorithm}
	\label{alg:2}
	\begin{algorithmic}[l]
		\REQUIRE Network $N$, weight matrix $W$, mask matrix $C$, data batch $B$, threshold $t$
		\IF{full growth}
		\STATE {set all elements in $C$ to 1}
		\ELSIF{gradient-based growth}
		\STATE {forward propagation through $N$ using data $B$ and then back propagation}
		\STATE {compute $g_{ij} = \left|\frac{\partial L}{\partial u_j}x_i\right|$}
		\STATE {For $g_{ij} > t$, set $c_{ij} = 1, w_{ij}=0 $ }
		\ENDIF
		\ENSURE Modified weight matrix $W$ and mask matrix $C$
		
	\end{algorithmic}
\end{algorithm}

\end{itemize}

\subsubsection{Neuron Growth}
Neuron growth adds new neurons to the network, thus gradually increases the network
size. 
Algorithm~\ref{alg:3} shows the process of neuron growth. 
By drawing an analogy from biological cell division, neuron
growth can be achieved by duplicating an existing neuron. 
To do this, we process a large batch of data through the network and compute the activation value of the neurons in the architecture. 
We choose the neurons with the highest activation values for duplication. 
To break the symmetry,
random noise is added to the weights of all the connections related to this
newly added neuron. 

\begin{algorithm}[h]
	\caption{Neuron growth algorithm}
	\label{alg:3}
	\begin{algorithmic}[l]
		\REQUIRE Network $N$, weight matrix $W$, mask matrix $C$, data batch $B$, a
candidate neuron $n_j$ to be added in
		\STATE {forward propagation through $N$ using data $B$}
		\STATE {$i = argmax~u_i$}
		\STATE {$c_{j\cdot} = c_{i\cdot}, c_{{\cdot}j} = c_{{\cdot}i}$}
		\STATE {$w_{j\cdot} = w_{i\cdot} + noise, w_{{\cdot}j} = w_{{\cdot}i} + noise$}
		
		\ENSURE Modified weight matrix $W$ and mask matrix $C$
		
	\end{algorithmic}
\end{algorithm}

\subsubsection{Connection Pruning}
Connection pruning disconnects previously connected neurons and reduces
the number of network parameters. If all connections associated with a neuron
are pruned, then the neuron is removed from the network. We adopt a widely-used method 
\cite{han2015learning,dai2019nest} to prune connections with small magnitude, as 
shown in Algorithm~\ref{alg:4}. The rationale
behind it is that since small weights have a relatively small influence on the
network, DNN performance can be restored through retraining after pruning.

\begin{algorithm}[h]
	\caption{Connection pruning algorithm}
	\label{alg:4}
	\begin{algorithmic}[l]
		\REQUIRE Weight matrix $W$, mask matrix $C$, threshold $t$
		\FORALL {$w_{ij}$}
		\IF{$\left| w_{ij} \right| < t$}
		\STATE {$c_{ij} = 0$}
		\ENDIF
		\ENDFOR
		\ENSURE Modified weight matrix $W$ and mask matrix $C$
		
	\end{algorithmic}
\end{algorithm} 

\subsection{Training Schemes}

In practice, depending on how the initial network architecture $A_{init}$ and basic operations in 
Step (a) of Algorithm \ref{alg:scann} are chosen, we adopt three training schemes in our
experiments, as explained next.  These training schemes enable us to synthesize different 
architectures with various structures, and yield compact and accurate models for each dataset. 

\subsubsection{Scheme A}
Scheme A is a constructive approach, where we start with a tiny network, and
gradually increase the network size.  
This can be achieved by performing
connection and neuron growth more often than connection pruning or carefully
selecting the growth and pruning rates, such that each growth operation grows
a large number of connections and neurons, while each pruning operation prunes
a small number of connections. 

To implement this scheme, we specify the initial number of hidden neurons (the minimum number of hidden neurons) in the architecture, as well as the maximum allowed number of hidden neurons in the final model. 
This scheme starts from the initial small number of hidden neurons, and applies connection 
growth, neuron growth, and connection pruning in this order. 
The neuron growth phase each time adds 
a certain number of neurons to the architecture (e.g., 5 or 10 neurons). 
In the connection 
growth process, we use gradient-based growth to add a certain percentile top connections (e.g. top $80\%$) to 
the network.  
Connection pruning is used to prune the network after each growth phase. 

\subsubsection{Scheme B}
Scheme B is a destructive approach, where we start with a large
network and make the network smaller by iteratively pruning connections.
One approach for accomplishing this \cite{han2015learning,dai2019nest} is based on iteratively 
pruning a small number of connections and then training the weights.  
This gradually
reduces the network size and finally results in a small network after many iterations.
We use a different method in Scheme B.  Rather than pruning the network gradually, we prune the 
network aggressively to a tiny size.  However, to recover the performance, we repeatedly prune 
the network and then grow the network back, rather than just perform gradual pruning and retraining. 

To implement this scheme, we start with a network architecture with a large number of hidden neurons. We consider the initial point as the maximum allowed number of hidden neurons in the architecture. 
We apply iterative gradient-based connection growth and magnitude-based connection pruning to train 
both the architecture and weights. For the connection growth process, we use the gradient-based 
growth to add a certain top percentile (e.g., $70\%$ to $90\%$) connections to the network. 
Subsequently, we use aggressive connection pruning to reduce the number of connections drastically. 
In addition, we train the architecture for $10$ to $20$ epochs after applying each
architecture-changing operation. We perform these operations for several ($5$-$10$) iterations. 

\subsubsection{Scheme C}
Similar to Scheme B, Scheme C is also a destructive approach.  The main difference is the use of 
MLP architectures in Scheme C.  This can be achieved by adjusting the connection growth algorithm 
to only allow connections between adjacent layers and not allow skipped connections. 
Scheme C can be viewed as an iterative version of the dense-sparse-dense technique proposed in 
\cite{han2016dsd}.

To implement Scheme C, we start with an FC MLP architecture and apply connection pruning to 
drastically reduce the number of connections in the network. Then, in several iterations, we 
apply full growth to recover all the connections in the network, followed by connection pruning 
to reduce network size. 

\begin{figure}
	\centering
	\subfloat[]{\includegraphics[width=1\linewidth]{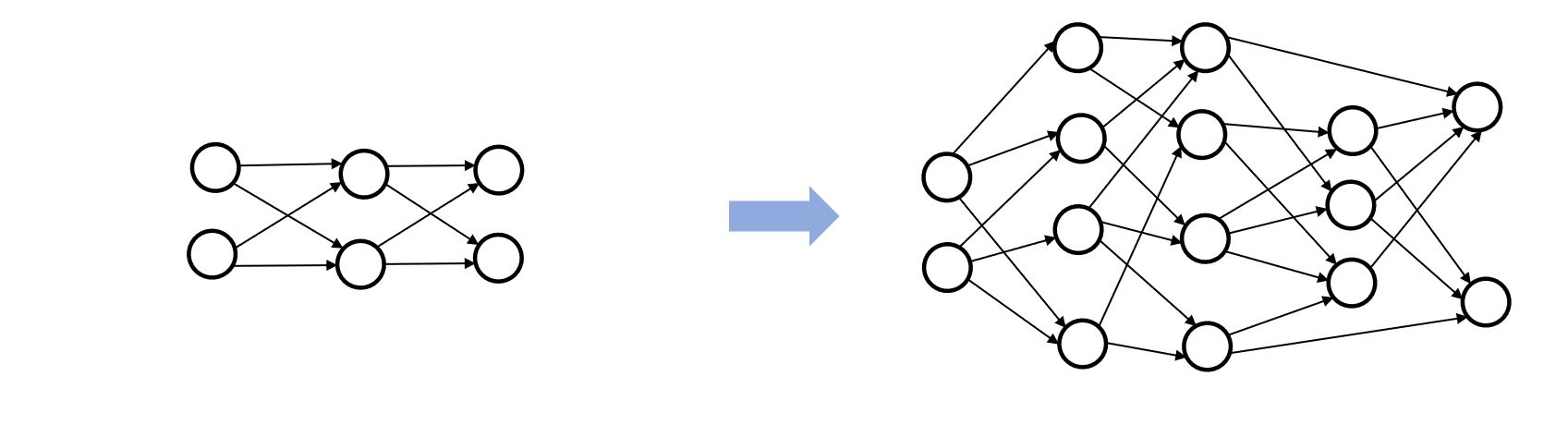}}
	\label{fig:Ng1}
	\hfil
	\subfloat[]{\includegraphics[width=1\linewidth]{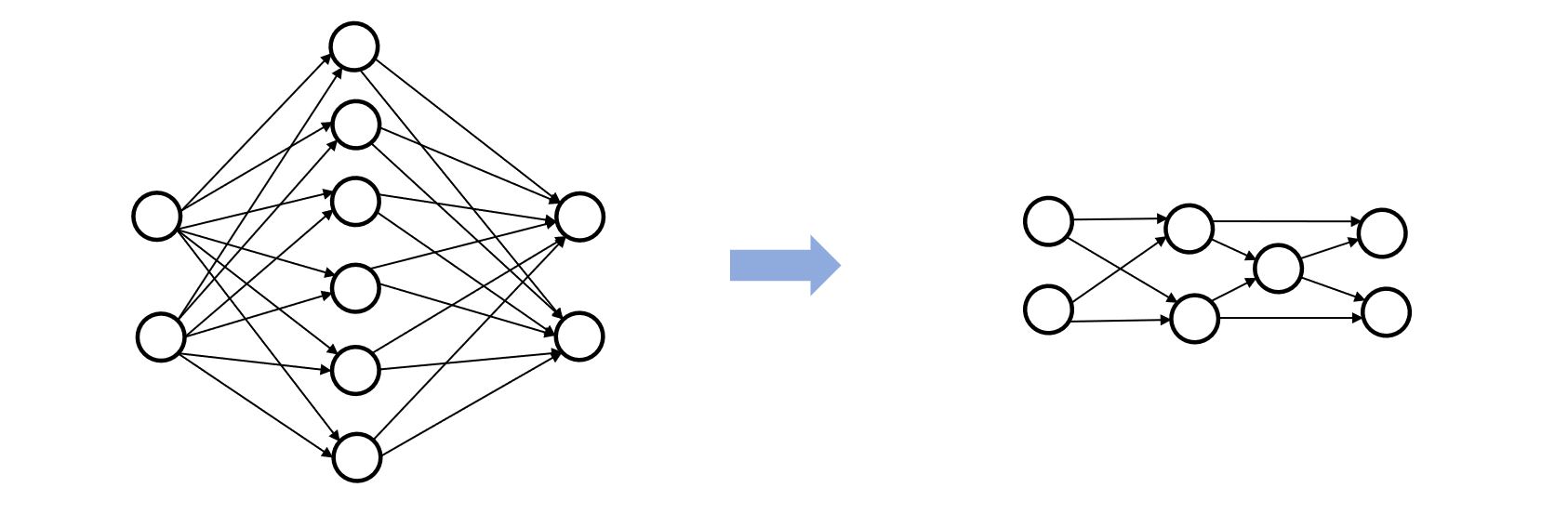}}
	\label{fig:Ng2}
	\subfloat[]{\includegraphics[width=1\linewidth]{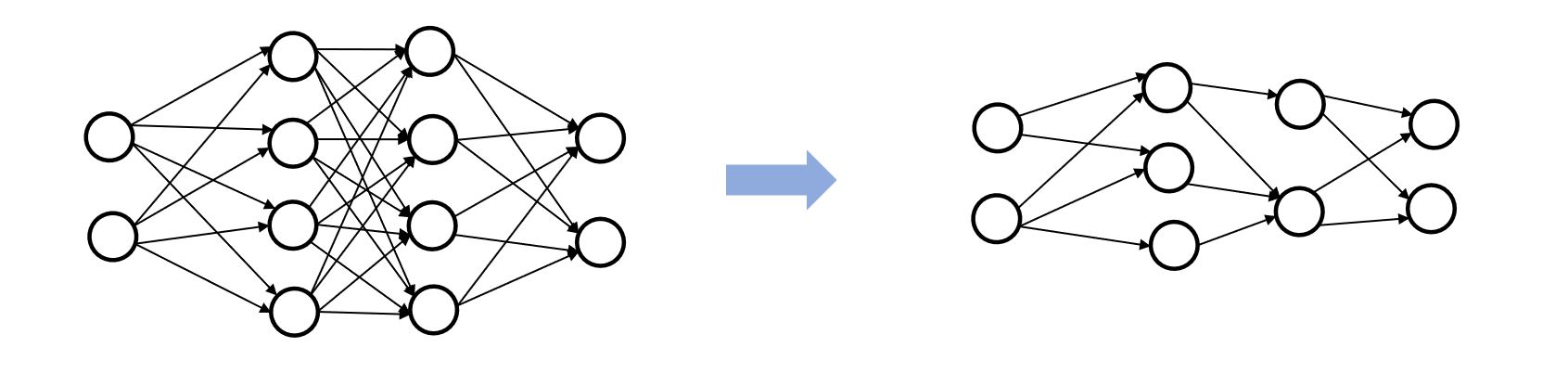}}
	\label{fig:Ng3}
	\caption{Illustration of the three training schemes. Shown are examples of
		initial and final architectures: (a) Scheme A, (b) Scheme B, and (c) Scheme C.}
	\label{schemes}
\end{figure}

Fig.~\ref{schemes} shows examples of the initial and final architectures
for each scheme. Both Schemes A and B evolve general feed-forward
architectures, thus allowing network depth to be learned during training.
Scheme C evolves an MLP structure, thus keeping the number of layers fixed.

\section{Experimental Results}
\label{sect:expresults}
In this section, we evaluate the performance of DR+SCANN and SCANN on nine small to medium size 
datasets, as well as on MNIST and ImageNet datasets.  
Table \ref{tab:characteristics} shows the characteristics of the nine datasets. 
For such non-image datasets, we compare our synthesized DNN model with the FC DNN architecture
that performs the best on the validation set.  For the MNIST and ImageNet datasets, we compare our 
synthesized models with various well-known benchmark architectures. 

The evaluation results are divided into two parts.  Section \ref{sect:results-first} discusses 
results obtained by SCANN when applied to image datasets: MNIST and ImageNet.  As we will see, 
SCANN generates neural networks with similar classification accuracy relative to well-known 
architectures, but with much fewer parameters and FLOPs.

In Section \ref{sect:results-second}, we present experimental resuls for DR, SCANN, and DR+SCANN 
methodologies, on nine non-image datasets.  We demonstrate that the DNNs
generated by SCANN and DR+SCANN are very compact and energy-efficient while maintaining performance. 
These results open up opportunities to use SCANN-generated DNNs in energy-constrained edge devices 
and IoT sensors.

\begin{table*}[htbp]
\caption{Characteristics of the datasets}
\label{tab:characteristics}
\centering
\begin{tabular}{lccccc}
\toprule
Dataset                                     & Training Set & Validation Set & Test Set & Features & Classes \\ \midrule
Sensorless Drive Diagnosis                  & $40509$      & $9000$         & $9000$   & $48$     & $11$   \\
Human Activity Recognition (HAR)            & $5881$       & $1471$         & $2947$   & $561$    & $6$     \\
Musk v$2$                                     & $4100$       & $1000$         & $1974$   & $166$    & $2$     \\
Pen-Based Recognition of Handwritten Digits & $5995$       & $1499$         & $3498$   & $16$     & $10$    \\
Landsat Satellite Image                     & $3104$       & $1331$         & $2000$   & $36$     & $6$     \\
Letter Recognition                          & $10500$      & $4500$         & $5000$   & $16$     & $26$    \\
Epileptic Seizure Recognition               & $6560$       & $1620$         & $3320$   & $178$    & $2$     \\
Smartphone Human Activity Recognition       & $6121$       & $153$          & $3277$   & $561$    & $12$    \\
DNA                                         & $1400$       & $600$          & $1186$   & $180$    & $3$     \\
\bottomrule
\end{tabular}
\end{table*}

\subsection{Experiments with MNIST and ImageNet}
\label{sect:results-first}
MNIST is a well-studied dataset of handwritten digits. It contains 60000 training images and 10000 
test images. We set aside 10000 images from the training set as the validation set.  
We first target the Lenet-5 Caffe model \cite{lecun1998gradient}.
The LeNet-5 Caffe model contains two convolutional layers with 20 and 50
filters, and also one FC hidden layer with 500 neurons. 

We use the stochastic gradient descent (SGD) optimizer
with a learning rate of 0.03, momentum of 0.9, and weight decay of 1e-4. 
For schemes A and B, the feed-forward part of the network is learnt by SCANN, while the convolutional 
part of the architecture is kept the same.  For scheme A, we start from 300 hidden neurons in the 
hidden layer, and set the maximum number of neurons to 500. 
At the beginning, we use connection pruning to prune 95 percent of the connections in the network. 
Subsequently, we apply connection growth, neuron growth, and connection pruning in several iterations.
The neuron growth operation duplicates the five neurons in the architecture with the highest 
activation values.  The connection growth activates
35 percent of all connections and connection pruning prunes 25 percent of the existing connections. 
In Scheme B, the best results correspond to 400 hidden neurons in the feed-forward part. We iteratively
perform a sequence of connection pruning such that 19.3K connections are left in the architecture,
and connection growth such that 90 percent of all connections are restored.
In Scheme C, we start the feed-forward part of the network with the FC part of the baseline 
architecture.  We iteratively prune the network to its final number of parameters and then 
use connection growth to restore all connections.

Table \ref{tab:b} summarizes the results. The baseline error rate is 0.72\% with 430.5K parameters. 
The most compressed model generated by SCANN contains only 9.3K parameters (with a compression
ratio of 46.3$\times$ over the baseline), achieving the same 0.72\% error rate when using Scheme C. 
Scheme A obtains the best error rate of 0.68\%, however, with a lower compression ratio of 
2.3$\times$.  For a fair comparison, we implement the method given in \cite{han2015learning} on 
the same data split.

\begin{table*}[t]
	\caption{Comparison of different methods on the LeNet-5 Caffe model}
	\label{tab:b}
	\centering
	\begin{tabular}{lcccccc}
		\toprule
		Methods     & Error rate &  \#Params &   Compression ratio \\
		\midrule
		Baseline    & 0.72\% & 430.5K & 1.0$\times$ \\
		
		Network pruning \cite{han2015learning} & 0.77\% & 34.5K & 12.5$\times$ \\
		Scheme A & 0.68\% & 184.6K  & 2.3$\times$ & \\
		Scheme B & 0.72\% & 19.3K  & 22.3$\times$ & \\
		Scheme C & 0.72\% & 9.3K  & 46.3$\times$ & \\
		\bottomrule
	\end{tabular}
\end{table*}

\begin{figure}[]
	\centering
	\includegraphics[width=3.5in]{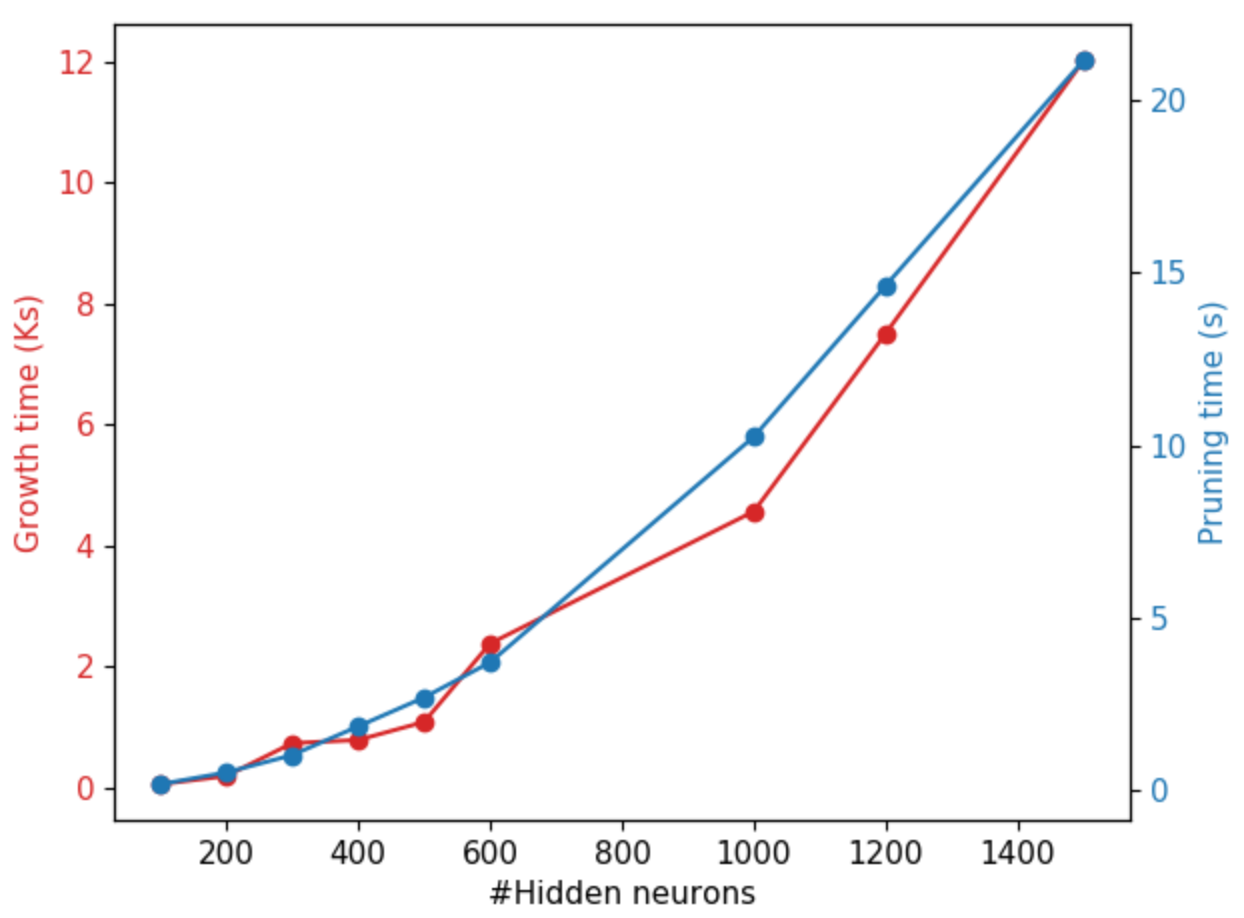}
	\caption{Growth and pruning time vs. the number of hidden neurons in the seed architecture for the MNIST dataset.}
	\label{fig:gp-time}
\end{figure}

Next, we study the impact of the seed architecture on GPU time (Nvidia Tesla P$100$) for growth and 
pruning operations on the MNIST dataset.  Fig.~\ref{fig:gp-time} demonstrates this trend for 
different numbers of maximum hidden neurons in the architecture (Scheme B was used in this case).  
The growth operation is more computationally intensive than the pruning operation. 
This is because while magnitude-based pruning only needs the forward pass, gradient-based 
growth needs both forward and backward passes on the network.  In addition, as the number of hidden 
neurons in the architecture increases, the GPU time of both operations increases, as expected.

We now use the feed-forward architecture proposed by Ciresan et 
al.~\cite{journals/corr/abs-1003-0358} as the baseline architecture for SCANN synthesis. 
This architecture has six layers with 2500, 2000, 1500, 1000, 500, and 10 neurons, respectively.  
As shown in Table \ref{tab:mnist-ff}, this baseline reduces the error rate to just 0.35\% through a 
dramatic increase in the number of parameters to 12.0M.  It represents the state-of-the-art test 
accuracy for an FC DNN on the MNIST benchmark. It consumes 23.9M FLOPs.
Thus, this network is computationally intensive and has significant memory requirements. 
We use this architecture as the starting point for SCANN Scheme C.  We use the SGD optimizer with 
an initial learning rate of $1$e-$3$ and gradually decrease it to $1$e-$6$.
We use connection pruning to remove $95$ percent of the connections in the network, and connection growth to restore all the connections. 
Through iterative growth and 
pruning, we are able to synthesize a much more compact architecture. 
We are able to reduce the number of parameters by $20.0\times$ and computational cost by 
$19.9\times$ with only a $0.02\%$ increase in error rate. 

\begin{table}[t]
	\caption{Further comparisons on MNIST}
	\label{tab:mnist-ff}
	\centering
	\begin{tabular}{lcccc}
		\toprule
		Methods     & Error rate &  \#Params & FLOPs \\
		\midrule
		Ciresan et al \cite{journals/corr/abs-1003-0358} & $0.35\%$ & 12.0M  & 23.9M \\
		Scheme C & $0.37\%$ & 0.6M (20.0$\times$)  & 1.2M (19.9$\times$)  \\
		\bottomrule
	\end{tabular}
\end{table}

To demonstrate the applicability of SCANN to different architectures and on different datasets of 
various sizes, we also use SCANN to synthesize DNNs for the ImageNet dataset \cite{deng2009imagenet}.
Table~\ref{tab:res-imagenet} shows the results of our experiments.  For the ImageNet experiments, we 
use the SGD optimizer with an initial learning rate of 0.05 and gradually decrease it to 1$e$-4. 
The weight decay is set to $4$e-$5$.
We initialize the grow-and-prune process with VGG-16 \cite{DBLP:journals/corr/SimonyanZ14a} and 
MobileNetV2 \cite{sandler2018mobilenetv2} architectures.  VGG-16 consists of 13 convolutional layers, 
5 max-pooling layers, and 3 FC layers.  We use SCANN to optimize the FC layers, where
most of the parameters reside, to learn the connections and weights in the training process.  
VGG-16 consists of 138.4M parameters.  Its FC layers contribute to 123.6M parameters of the 
architecture.  Thus, reducing the number of parameters in the FC layers can have a significant impact 
on the model size.  Our best result is obtained using SCANN Scheme B.
We set the number of hidden neurons in the architecture to 4000.  Initially, we prune 95 percent of 
the connections. Next, we use connection growth to grow 30 percent of the connections, followed by 
connection pruning to leave only 2.5M connections in the feed-forward part of the architecture.
As a result, SCANN reduces the number of parameters to $17.2$M for an 8.0$\times$ reduction. 
In addition, SCANN reduces the top-1 error rate by 1.7\% to $26.7\%$.  However, since most of the 
computational cost of a CNN architecture is in its convolution operations, SCANN is not able to 
reduce the FLOPs much. 

MobileNetV2 is an architecture optimized for mobile devices.  Hence, it has reduced computational 
cost.  Its FC layer contains $37\%$ of all its parameters. Keeping the rest of the architecture
fixed, we use SCANN Scheme C to optimize the FC layer.  We use connection pruning to remove 800K 
connections in the FC layer. Subsequently, we use connection growth to restore all the connections. 
Using iterative growth and pruning, we can reduce the number of
parameters by 1.3$\times$ at the cost of a slight 0.2\% increase in the error rate.

\begin{table}[t]
	\caption{Comparison of different methods on the ImageNet dataset}
	\label{tab:res-imagenet}
	\centering
	\begin{tabular}{lcccc}
		\toprule
		Methods     & Top-1 error rate  &  \#Params & FLOPs \\
		\midrule
		VGG-16 \cite{DBLP:journals/corr/SimonyanZ14a} & $28.4\%$ & 138.4M  & 30.9B \\
		Our VGG-16 & $26.7\%$ & $17.2$M ($8.0\times$) & $30.6$B \\ \midrule
		MobileNetV2 \cite{sandler2018mobilenetv2} & $28.0\%$ & 3.4M  & 300M  \\
		Our MobileNetV2 & $28.2\%$ & 2.6M ($1.3\times$) & 298M \\
		\bottomrule
	\end{tabular}
\end{table}

While RL-based architecture search approaches, such as NASNet \cite{DBLP:conf/cvpr/ZophVSL18}, 
consume around 2000 GPU days for the ImageNet dataset, SCANN requires around 20 GPU days for
optimizing the FC layers of a given architecture.

\subsection{Experiments with Other Datasets}
\label{sect:results-second}
To demonstrate the capability of SCANN and DR+SCANN for synthesizing accurate and compact neural 
network models for various non-image datasets, we experiment with nine datasets from the UCI machine 
learning repository \cite{Dua:2017} and Statlog collection \cite{Michie:1995:MLN:212782}. 
Next, we present evaluation results on these datasets.

SCANN experiments are based on the Adam optimizer with a learning rate of $0.01$ and weight decay 
of $1$e-$3$.  We compare results obtained by DR+SCANN with those obtained by only applying SCANN and 
also DR without using SCANN in a secondary compression step.  Table \ref{smallaccuracy2} shows the 
classification test accuracy obtained.  The MLP column shows the accuracy of the best MLP baseline 
found with the help of the validation set.  For all the other 
methods, we present two columns, the left of which shows the highest accuracy (H.A.) achieved
whereas the right one shows the result for the most compressed (M.C.) network. Furthermore, 
in the DR columns, the DR method employed is shown in parentheses.  In the DR columns, whereas 
the M.C. and H.A. columns employ the same DR method, they may use different DR ratios. 
Table \ref{smallcompression2} shows the number of parameters in the network for the corresponding 
columns in Table \ref{smallaccuracy2}. 

SCANN-generated networks show improved accuracy for six of the nine datasets, as compared to 
the MLP baseline. The accuracy increase is between $0.41\%$ to $9.43\%$. These results correspond to 
networks that are $1.2\times$ to $42.4\times$ smaller than the baseline architecture. Furthermore, 
DR+SCANN shows improvements on the highest classification accuracy in five out of the nine
datasets, as compared to SCANN-generated results.
In addition, SCANN yields DNNs that achieve the baseline accuracy with fewer parameters in
seven out of the nine datasets. For these datasets, the results show a parameter compression ratio 
between $1.5\times$ to $317.4\times$.  Moreover, as shown in Tables \ref{smallaccuracy2} and 
\ref{smallcompression2}, combining DR with SCANN helps achieve higher 
compression ratios. For these seven datasets, DR+SCANN can meet the baseline accuracy with 
a $28.0\times$ to $5078.7\times$ smaller network. This shows a significant improvement over the 
compression ratio achievable by just using SCANN. 

We also report the performance of applying DR without the benefit of the SCANN synthesis step. While 
these results show improvements, DR+SCANN can be seen to have much more compression power, 
relative to when DR and SCANN are used separately.  This points to a synergy between DR and SCANN.

\begin{table*}[t]
\caption{Test accuracy comparison}
\label{smallaccuracy2}
\centering
\begin{tabular}{lccccccc} \toprule
Dataset          & MLP       & DR (H.A.)                 & DR (M.C.)                & SCANN (H.A.)    & SCANN (M.C.)   & DR+SCANN (H.A.)& DR+SCANN (M.C.) \\ \midrule
SenDrive & $93.53\%$ & $99.07\%$ (FA)     & $97.99\%$ (FA)     & $97.10\%$  & $93.63\%$ &
$99.34\%$  & $94.20\%$    \\
HAR              & $95.01\%$ & $95.04\%$ (ICA)    & $95.04\%$ (ICA)    & $95.52\%$ & $95.52\%$ & $95.28\%$  & $95.08\%$   \\
Musk             & $98.68\%$ & $98.83\%$ (FA)     & $98.78\%$ (FA)     & $99.09\%$ & $98.83\%$ & $98.08\%$  & $98.08\%$   \\
Pendigits        & $97.22\%$ & $97.51\%$ (Isomap) & $97.39\%$ (Isomap) & $97.22\%$ & $97.22\%$ & $97.93\%$  & $97.65\%$   \\
SatIm            & $91.30\%$  & $91.10\%$ (PCA)     & $91.10\%$ (PCA)     & $90.10\%$  &
$90.10\%$  & $89.40\%$   & $89.40\%$    \\
Letter           & $95.24\%$ & $94.92\%$ (PCA)    & $94.92\%$ (PCA)    & $92.60\%$ & $92.60\%$ & $92.70\%$  & $92.70\%$   \\
Seizure          & $87.53\%$ & $97.50\%$ (FA)     & $95.42\%$ (FA)     & $96.96\%$ & $96.23\%$ & $97.62\%$  & $95.72\%$ \\
SHAR             & $90.66\%$ & $94.44\%$ (RP)     & $90.69\%$ (RP)     & $93.78\%$ & $90.93\%$ & $94.84\%$  & $90.93\%$   \\
DNA              & $94.86\%$ & $94.69\%$ (FA)     & $94.69\%$ (FA)     & $95.86\%$ & $95.36\%$ & $93.76\%$  & $93.76\%$  \\ \bottomrule
\end{tabular}
\end{table*}

\begin{table*}[t]
\caption{Neural network parameter comparison}
\label{smallcompression2}
\centering
\begin{tabular}{lccccccc} \toprule
Dataset          & MLP                   & DR (H.A.)                   & DR (M.C.)                    & SCANN (H.A.)                 & SCANN (M.C.)                  & DR+SCANN (H.A.)               & DR+SCANN (M.C.)             \\ \midrule
SenDrive & $56.9$k ($1 \times$)  & $2.6$k ($21.9\times$)  & $1140$ ($49.9\times$)  & $10.0$k ($5.7\times$)   & $750$ ($75.9\times$)    & $2.2$k ($25.9\times$)  & $200$ ($284.5\times$) \\
HAR              & $212.0$k ($1 \times$) & $108.4$k ($1.9\times$) & $108.4$k ($1.9\times$) & $5.0$k ($42.4\times$) & $5.0$k ($42.4\times$)   & $1.0$k ($212\times$)   & $750$ ($282.7\times$) \\
Musk             & $55.8$k ($1 \times$)  & $17.3$k ($3.2\times$)  & $15.5$k ($3.6\times$)  &
$22.0$k ($2.5\times$) & $20.0$k ($2.8\times$)   & $600$ ($93.0\times$)     & $600$ ($93.0\times$)    \\
Pendigits        & $4.9$k ($1 \times$)   & $780$ ($6.3\times$)    & $671$ ($7.3\times$)    &
$3.2$k ($1.5\times$)  & $3.2$k ($1.5\times$)    & $400$ ($12.2\times$)   & $175$ ($28.0\times$)    \\
SatIm            & $3.8$k ($1 \times$)   & $1.1$k ($3.4\times$)   & $1.1$k ($3.4\times$)   & $3.2$k ($1.5\times$)  & $3.2$k ($1.5\times$)    & $1.0$k ($3.8\times$)   & $1.0$k ($3.8\times$)  \\
Letter           & $4.4$k ($1 \times$)   & $3.8$k ($1.1\times$)   & $3.8$k ($1.1\times$)   & $3.8$k ($1.1\times$)  & $3.8$k ($1.1\times$)    & $3.7$k ($1.2\times$)   & $3.7$k ($1.2\times$)  \\
Seizure          & $380.9$k ($1 \times$) & $10.5$k ($36.3\times$) & $616$ ($618.3\times$)  &
$3.0$k ($127.0\times$)  & $1.2$k ($317.4 \times$) & $1.8$k ($211.6\times$) & $75$ ($5078.7\times$) \\
SHAR             & $214.0$k ($1 \times$)   & $127.1$k ($1.7\times$) & $3.7$k ($57.8\times$)  &
$10.0$k ($21.4\times$)  & $800$ ($267.5\times$)   & $1.0$k ($214.0\times$)   & $500$ ($428.0\times$)   \\
DNA              & $24.6$k ($1 \times$)  & $22.9$k ($1.1\times$)  & $22.9$k ($1.1\times$)  &
$20.0$k ($1.2\times$)   & $200$ ($123.0\times$)     & $300$ ($82.0\times$)     & $300$ ($82.0\times$) \\ \bottomrule  
\end{tabular}
\end{table*}

Although classification performance is of great importance, in applications where computing resources 
are limited, e.g., in battery-operated devices, energy efficiency might be a very
important concern. Thus, the energy performance of the models should also be taken into 
consideration in such cases. To evaluate the energy performance, we use the energy
analysis method proposed in \cite{akmandor2018simultaneously}, where the
energy consumption for inference is calculated based on the number of
multiply-accumulate (MAC) and comparison operations and the number of SRAM
accesses. For example, a multiplication of two matrices of size $M \times N$
and $N \times K$ would require $(M \cdot N \cdot K)$ MAC operations and
$(2 \cdot M \cdot N \cdot K)$ SRAM accesses. In this energy model, a single
MAC operation, SRAM access, and comparison operation implemented in a 130$nm$
CMOS process (which may be an appropriate technology for many IoT sensors) consumes 11.8 $pJ$, 
34.6 $pJ$, and 6.16 $fJ$, respectively.  Table \ref{tab:j} shows the energy consumption estimates 
per inference for the models presented in Tables \ref{smallaccuracy2} and \ref{smallcompression2}. 
Note that energy consumption does not include dataset DR. However, some of the
DR methods, like RP, just require a single matrix-vector multiplication. Hence, such
methods do not have much energy overhead.

As can be seen, SCANN models are significantly more energy-efficient compared to FC baselines. 
In addition, DR+SCANN can be seen to have the best overall energy performance. Except for 
the Letter dataset (for which the energy reduction is only 17 percent), the compact DNNs generated by 
DR+SCANN consume one to four orders of magnitude less energy than the baseline MLP models. 
Thus, SCANN and DR+SCANN synthesis methodologies are suitable for heavily energy-constrained devices, 
such as IoT sensors.

\begin{table*}[t]
\caption{Inference energy consumption comparison ($J$)}
\label{tab:j}
\centering
\begin{tabular}{lccccccc} \toprule
Dataset          & MLP                   & DR (H.A.)                   & DR (M.C.)                    & SCANN (H.A.)                 & SCANN (M.C.)                  & DR+SCANN (H.A.)               & DR+SCANN (M.C.)             \\ 
\midrule
SenDrive & $4.6$e-$6$  & $2.1$e-$7$   & $8.9$e-$8$ & $8.1$e-$7$   & $6.1$e-$8$ & $1.8$e-$7$ & $1.6$e-$8$ \\
HAR              & $17.2$e-$6$ & $8.8$e-$6$ & $8.8$e-$6$ & $4.0$e-$7$   & $4.0$e-$7$ & $8.1$e-$8$ & $6.1$e-$8$ \\
Musk             & $4.5$e-$6$  & $1.4$e-$6$   & $1.2$e-$6$ & $1.8$e-$6$ & $1.6$e-$6$ & $4.9$e-$8$ & $4.9$e-$8$ \\
Pendigits        & $4.0$e-$7$  & $6.3$e-$8$   & $5.4$e-$8$ & $2.6$e-$7$   & $2.6$e-$7$ & $3.2$e-$8$ & $1.4$e-$8$ \\
SatIm            & $3.1$e-$7$  & $8.9$e-$8$   & $8.9$e-$8$ & $2.6$e-$7$   & $2.6$e-$7$ & $8.1$e-$8$ & $8.1$e-$8$ \\
Letter           & $3.6$e-$7$  & $3.1$e-$7$   & $3.1$e-$7$ & $3.1$e-$7$   & $3.1$e-$7$ & $3.0$e-$7$ & $3.0$e-$7$ \\
Seizure          & $3.1$e-$5$  & $8.5$e-$7$   & $5.0$e-$8$ & $2.4$e-$7$   & $9.7$e-$8$ & $1.4$e-$7$ & $6.1$e-$9$ \\
SHAR             & $1.7$e-$5$  & $1.0$e-$5$   & $3.0$e-$7$ & $8.1$e-$7$   & $6.5$e-$8$ & $8.1$e-$8$ & $4.0$e-$8$ \\
DNA              & $2.0$e-$6$  & $1.8$e-$6$   & $1.8$e-$6$ & $1.6$e-$6$   & $1.6$e-$8$ & $2.4$e-$8$ & $2.4$e-$8$ \\ \bottomrule

\end{tabular}
\end{table*}

\section{Discussion}
\label{sect:discussion}

The advantages of SCANN are derived from its core benefit: the network
architecture is allowed to dynamically evolve during training. This benefit is
not directly available in several other existing automatic DNN architecture
synthesis techniques, such as the evolutionary and RL-based
approaches. In those methods, a new architecture, whether generated
through mutation and crossover in the evolutionary approach or from
the controller in the RL approach, needs to be fixed
during training and trained from scratch again when the architecture is
changed. However, human learning is incremental.  Our brain gradually changes
based on the presented stimuli.  For example, studies of the human neocortex
have shown that up to 40 percent of the synapses are rewired every day
\cite{hawkins2017intelligent}.  Hence, from this perspective, SCANN
takes inspiration from how the human brain evolves incrementally.
SCANN's dynamic rewiring is easily achieved through connection growth and pruning.

Comparisons between SCANN and DR+SCANN show that the latter results in a 
smaller network in nearly all the cases. This is due to preceding SCANN with
DR. By mapping data instances into lower dimensions, it reduces the number of 
neurons in each layer of the DNN, without degrading performance. This enables SCANN to
start with a significantly smaller DNN.  However, a limitation of SCANN is that it can only evolve 
feed-forward networks. How to extend SCANN to the convolutional layers in CNNs and recurrent neural
networks is the focus of our future work.


\section{Conclusion}
\label{sect:conclusion}
In this article, we proposed a synthesis methodology that can generate compact
and accurate neural networks. It solves the problem of having to fix the
depth of the network during training that prior synthesis methods suffer from.
It is able to evolve an arbitrary feed-forward network architecture with
the help of three basic operations: connections growth, neuron growth, and
connection pruning. 
Furthermore, by combining DR with SCANN synthesis, we showed significant improvements in the 
network compression power of this framework.  Experiments on MNIST and ImageNet image datasets, and 
several other small to medium non-image datasets, showed that SCANN and DR+SCANN can provide a good 
tradeoff between accuracy, model compression, and energy efficiency in applications where computing 
resources are limited. 

\bibliographystyle{IEEEtran}
\bibliography{SCANN}

\vspace{-1.0cm}
\begin{IEEEbiography}[
{\includegraphics[width=1.0in,height=1.1in,clip,keepaspectratio]{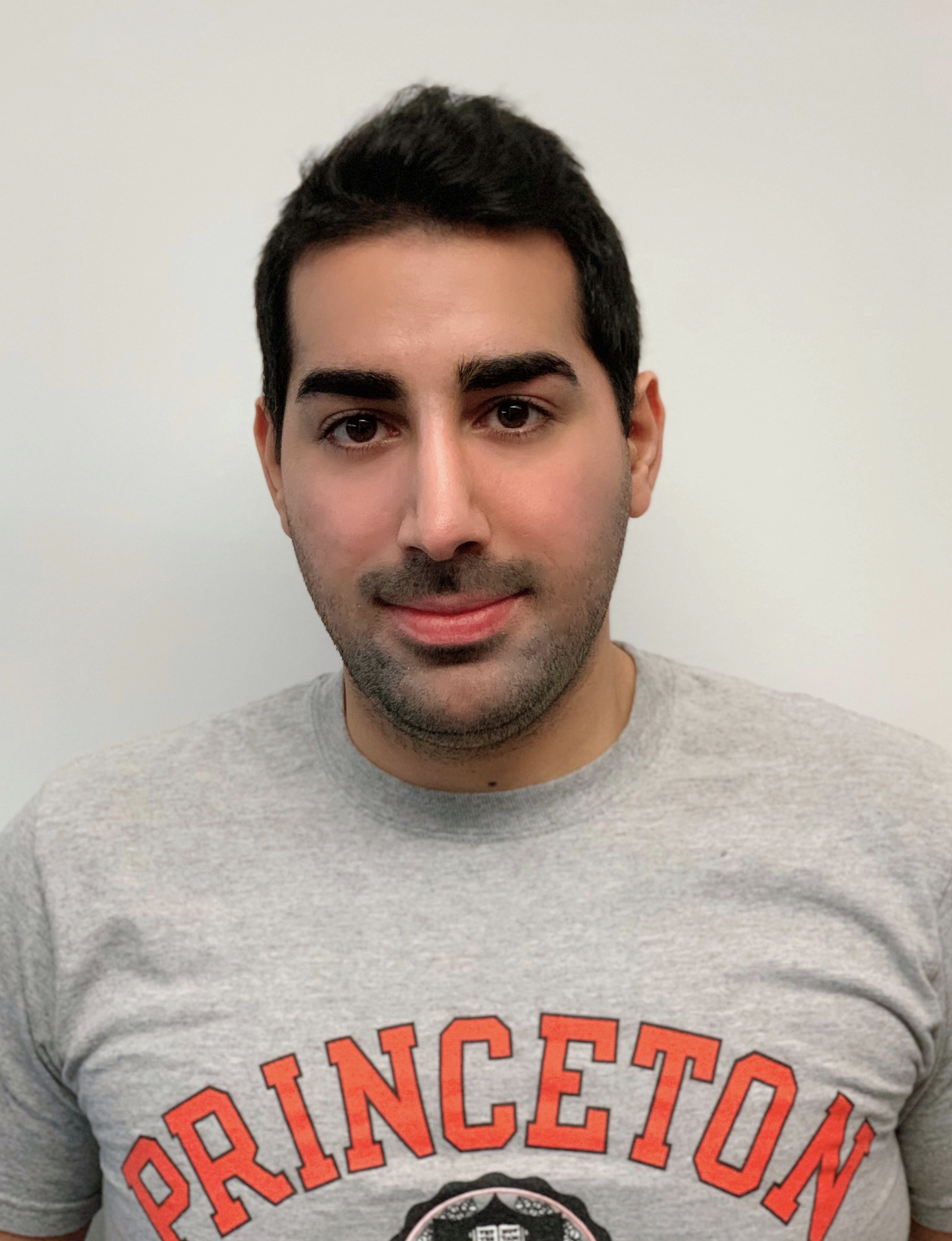}}]
{Shayan Hassantabar} received his B.S. degree in Electrical Engineering, Digital Systems focus, from Sharif University of Technology, Iran. He also received his M.Math. degree in Computer Science from University of Waterloo, Canada, and his M.A. degree in Electrical and Computer Engineering from Princeton University. He is pursuing the Ph.D. degree in Electrical and Computer Engineering at Princeton University. His research interests include automated neural network architecture synthesis, neural network compression, and smart healthcare.
\end{IEEEbiography}

\vspace{-1.0cm}
\begin{IEEEbiography}[
{\includegraphics[width=1.0in,height=1.1in,clip,keepaspectratio]{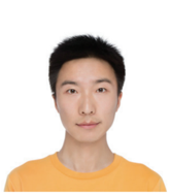}}]
{Zeyu Wang} received his B.S. degree in 2016 from Fudan University, Shanghai, China. He is currently pursuing his Ph.D. in the Electrical and Computer Engineering Department at Princeton University. His research interests include machine learning, artificial intelligence, neural network, deep learning and its applications.

\end{IEEEbiography}

\vspace{-1.0cm}
\begin{IEEEbiography}[
{\includegraphics[width=1.0in,height=1.1in,clip,keepaspectratio]{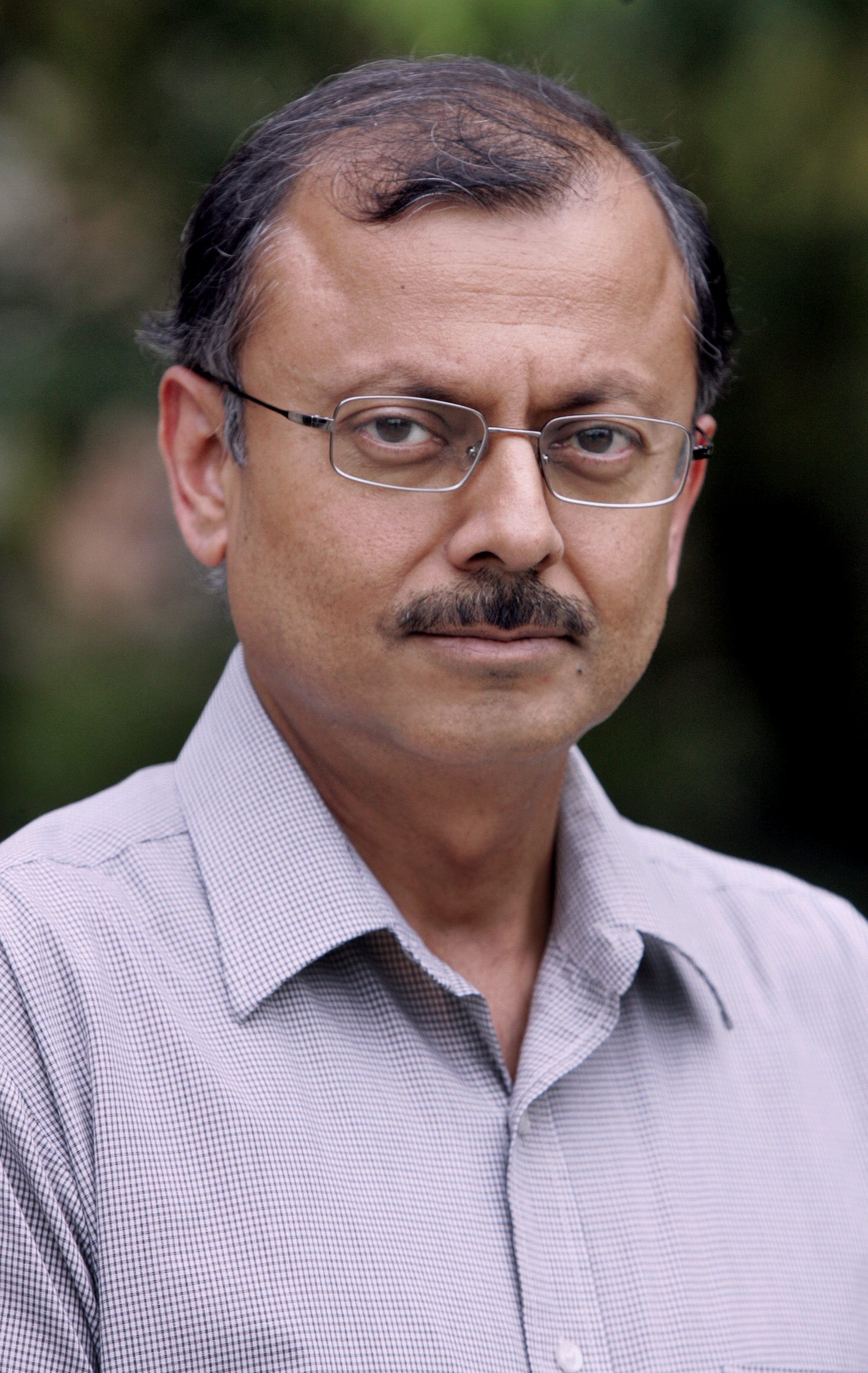}}]
{Niraj K. Jha} received his B.Tech. degree in Electronics and Electrical Communication Engineering from Indian Institute of Technology, Kharagpur, India in 1981 and Ph.D. degree in Electrical Engineering from University of Illinois at Urbana-Champaign, IL in 1985. He has been a faculty member of the Department of Electrical Engineering, Princeton University, since 1987. He is a Fellow of IEEE and ACM, and was given the Distinguished Alumnus Award by I.I.T., Kharagpur. He has received the Princeton Graduate Mentoring Award.

He has served as the Editor-in-Chief of IEEE Transactions on VLSI Systems and an Associate Editor of several other journals. He has co-authored five widely used books. His research has won 20 best paper awards or nominations and 21 patents.  His research interests include smart healthcare, cybersecurity, machine learning, and monolithic 3D IC design. He has given several keynote speeches in the areas of nanoelectronic design/test, smart healthcare, and cybersecurity.
\end{IEEEbiography}
\end{document}